\documentclass{article}

\PassOptionsToPackage{numbers, compress}{natbib}

\usepackage[preprint]{neurips_2026}

\usepackage[utf8]{inputenc} 
\usepackage[T1]{fontenc}    
\usepackage{hyperref}       
\usepackage{url}            
\usepackage{booktabs}       
\usepackage{amsfonts}       
\usepackage{nicefrac}       
\usepackage{microtype}      
\usepackage{xcolor}         

\usepackage{graphicx}
\usepackage{subcaption}
\usepackage{booktabs}
\usepackage{wrapfig}        
\usepackage{lipsum}

\definecolor{cvprblue}{rgb}{0.21,0.49,0.74}

\definecolor{textgreen}{rgb}{0.4980392156862745, 0.788235294117647, 0.4980392156862745}
\definecolor{ourmediumblue}{rgb}{0.21568627450980393,0.49411764705882355,0.7215686274509804}
\definecolor{ourmediumred}{rgb}{0.8941176470588236,0.10196078431372549,0.10980392156862745}
\definecolor{ourmediumgreen}{rgb}{0.30196078431372547,0.6862745098039216,0.2901960784313726}

\definecolor{ourlightred}{rgb}{0.984313725490196, 0.5019607843137255, 0.4470588235294118}
\definecolor{ourlightblue}{rgb}{0.5019607843137255, 0.6941176470588235, 0.8274509803921568}
\definecolor{ourlightcyan}{rgb}{0.5529411764705883, 0.8274509803921568, 0.7803921568627451}

\definecolor{darkspringgreen}{rgb}{0.09, 0.45, 0.27}

\definecolor{cornellred}{rgb}{0.7, 0.11, 0.11}
\definecolor{darkcerulean}{rgb}{0.03, 0.27, 0.49}
\definecolor{amaranth}{rgb}{0.9, 0.17, 0.31}
\definecolor{americanrose}{rgb}{1.0, 0.01, 0.24}

\definecolor{blue(ryb)}{rgb}{0.01, 0.28, 1.0}
\definecolor{cadmiumred}{rgb}{0.89, 0.0, 0.13}
\definecolor{darkorange}{rgb}{1.0, 0.55, 0.0}
\definecolor{flame}{rgb}{0.89, 0.35, 0.13}

\newcommand{\txtdeepblue}[1]{\textcolor{darkcerulean}{#1}}
\newcommand{\txtorange}[1]{\textcolor{flame}{#1}}

\newcommand{\projectname}{\emph{Flame3D}}
\newcommand{\projectnamecolor}{\textbf{\txtorange{Flame}3D}}

\newcommand{\benchmarkname}{\emph{Compose3D}}
\newcommand{\benchmarknamecolor}{\textbf{\txtdeepblue{Compose}}3D}

\newcommand{\website}{\href{https://www.open-flame.com/flame3d/}{\textcolor{blue}{open-flame.com/flame3d}}}

\newcommand{\posdiff}[1]{\textcolor{green!45!black}{\tiny $\Delta$+#1}}
\newcommand{\negdiff}[1]{\textcolor{red!70!black}{\tiny $\Delta$#1}}

\title{\projectnamecolor: Zero-shot Compositional Reasoning of \\ 3D Scenes with Agentic Language Models \\{\normalsize \website} \vspace{-0.5em}}

\author{%
  Sagar Bharadwaj \quad Ziyong Ma \quad Anurag Ghosh \quad Srinivasan Seshan \quad Anthony Rowe \\
  Carnegie Mellon University \\
  \texttt{\{skalasib, ziyongm, anuraggh, srini, agr\}@andrew.cmu.edu}
}

\begin{document}

\maketitle

\begin{figure}[h]
    \vspace{-2em}
    \centering
    \includegraphics[width=\linewidth]{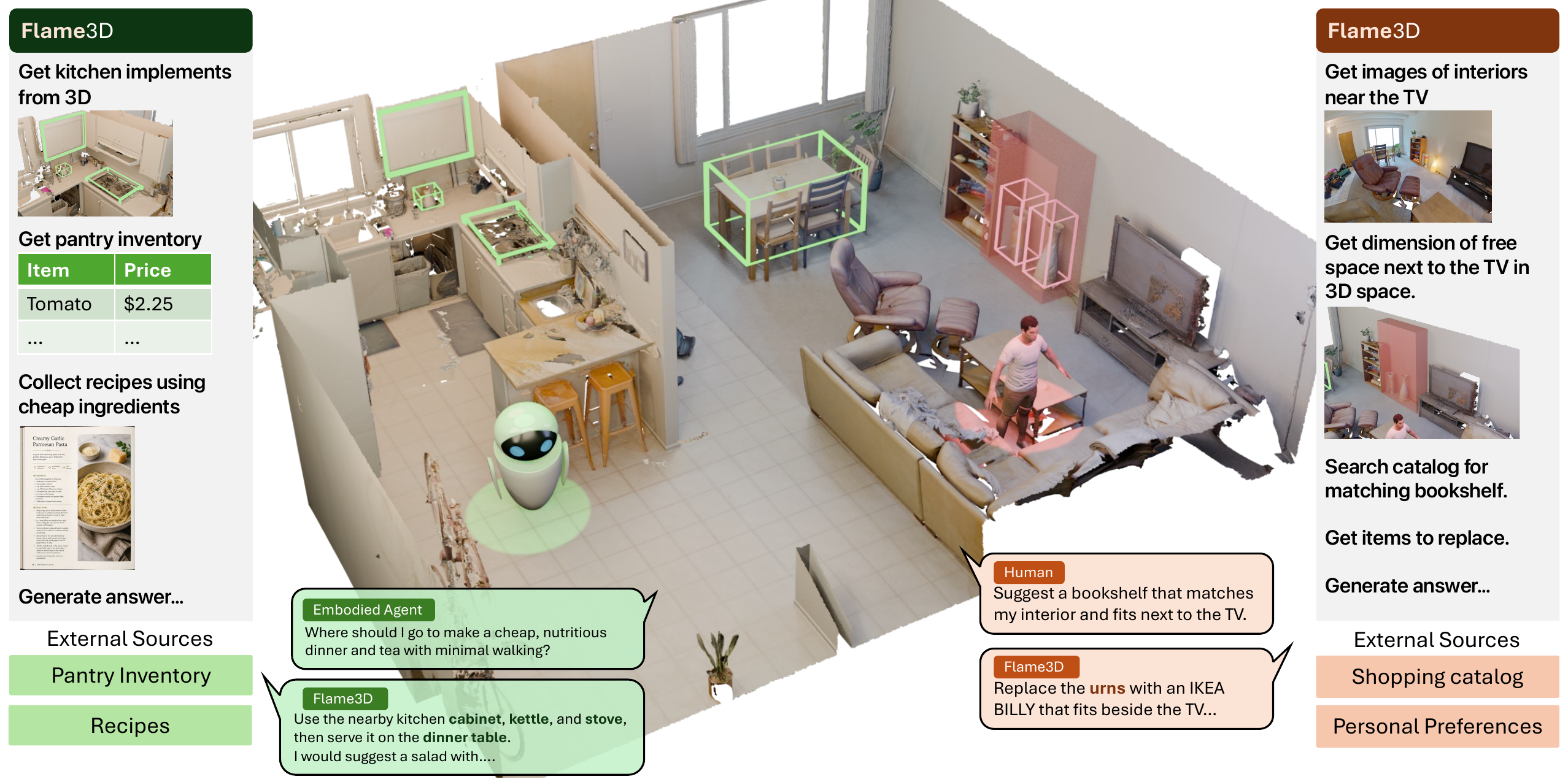}
    \caption{\projectnamecolor{} \textbf{answers compositional, multi-hop spatial queries about a 3D scene} by composing chain-of-thought inferences over a structured visual-textual scene memory and a systematically designed set of spatial and visual tools along with external knowledge. Qualitative examples are provided our website: \website{}}
    \label{fig:hero}
\end{figure}

\vspace{-0.5em}
\begin{abstract}
\vspace{-1em}
3D scene understanding spans reasoning about free space, object grounding, hypothetical object insertions, 
complex geometric relationships, and integrating all of these with external tools and data sources. 
Existing 3D understanding methods typically rely on large-scale 3D-language training or focus on object 
grounding and simple spatial relationships. We argue that the broad generalization that motivates 3D--language 
training can be achieved at inference time, without 3D-specific training. We propose \projectname{}\footnote{\textit{Flame} is an acronym for Flexible Language Abstractions for Memory-grounded Enquiry}, a 
training-free framework that represents scenes as editable visual-textual 3D memories and exposes them to an 
off-the-shelf MLLM through composable spatial tools. \projectname{} also lets the agent synthesize custom spatial 
programs at inference time, enabling open-ended reasoning over layouts, empty space, and objects not yet present 
in the scene. External data and corrections can be added to the memory without retraining.  In addition to showing competitive performance to finetuned 3D-LMM methods on ScanQA, we study multi-hop 3D reasoning capabilities of \projectname{} by evaluating it on a curated compositional spatial-reasoning benchmark, \benchmarkname{} . We find that fixed tools fall short and that the agent's ability to synthesize spatial 
operations at inference time is essential. These results invite the question: should future progress in 3D scene 
understanding focus on richer scene memories and expressive compositional abstractions?
\end{abstract}

\section{Introduction}
\label{sec:intro}

Consider the task shown in Figure~\ref{fig:hero}: ``Suggest a bookshelf that matches my interior and fits next to the TV''. Answering this query requires synthesizing diverse information: spatial relationships (`next to the TV'), physical layouts (`free space next to the TV'), visual aesthetics (`matches my interior'), and external knowledge (`suggest a bookshelf \textit{[from Ikea]}'). Practical 3D scene queries inherently demand compositional reasoning and frequently draw upon knowledge that resides outside the scene itself. Resolving these queries requires chaining spatial relations and layouts with semantic attributes, physical reasoning, and external context such as a shopping catalog or pantry inventory. Importantly, the reasoning involves not only grounding objects in space, but also reasoning about free spaces, for example, to accommodate objects not present in the scene. This motivates our central question: can we leverage the compositional reasoning capabilities of modern large language models (LLMs) in visual and textual domains and extend them to 3D scene reasoning?


Multiple lines of work tackle 3D scene understanding and grounding. Finetuned 3D large multimodal models (3D-LMMs)~\citep{hong20233d_3dllm,wang2023chat3d,mao2025spatialm,cheng2024spatialrgpt} treat 3D as a new modality aligned into the latent space of a large language model (LLM) through tokenized point clouds. However, this approach incurs significant costs such as expensive paired 3D--language data, limited compositional generalization beyond the training distribution, and the risk of catastrophic forgetting~\citep{kirkpatrick2017overcoming}. Recent work has further questioned whether point-cloud tokens help LLM spatial reasoning at all~\citep{zhang2025point} and whether finetuned 3D-LMMs actually understand 3D spatial relationships~\citep{ma20263real3dqa}. Alternatively, visual feature fields~\citep{jatavallabhula2023conceptfusion,kerr2023lerf,werby2024hierarchical} project pretrained embeddings, such as CLIP~\citep{pmlr_CLIP}, into 3D, producing representations whose semantics are frozen at construction time. This results in limited editability of 3D semantics, which is essential in practice.

A growing line of zero-shot work sidesteps training by feeding keyframes and viewpoints~\cite{taguchi2025spatialprompting, li2025see, zhang2024agent3dzero} or curated 3D operations~\cite{LLMGrounder, zantout2025sort3d, rana2023sayplan, linok2025beyond} to off-the-shelf multimodal large language models (MLLMs). However, these methods focus primarily on object grounding and object-centric reasoning. We argue that genuine 3D scene understanding is substantially broader. For example, it encompasses free-space reasoning to navigate or utilize empty volumes, hypothetical object insertions to imagine new spatial configurations, reasoning over complex geometry and layouts, and seamless external knowledge integration as seen in Figure~\ref{fig:hero}. 


We propose \projectname{}, a zero-shot framework that solves these challenges through two key mechanisms: compositionality of abstrations or tools over a structured representation, and meta tools for long-tail complex scenarios. Our framework constructs a structured visual-textual scene memory (\S\ref{sec:scene-memory}) which is then exposed to an off-the-shelf MLLM through a systematically designed set of spatial and visual tools (\S\ref{sec:abstractions}), as well as meta-tools that let the agent write executable code on the fly. An advantage of explicit representation is its inherent editability that allows updates to scene information without retraining or re-encoding. Furthermore, \projectname{} can easily accommodate external tools and sources of data (e.g., compliance, inventory) without architectural changes. \projectname{} requires no 3D-specific fine-tuning; as the underlying MLLMs improve, so does \projectname{}.

\projectname{} is competitive with finetuned 3D-LMMs, feature fields, and zero-shot methods on ScanQA~\citep{azuma2022scanqa}. To evaluate compositional reasoning, we create a new benchmark designed to prevent models from solely relying on learned language priors while answering questions seen on existing benchmarks~\cite{ma20263real3dqa}. Our evaluation reveals several key insights into agentic 3D reasoning. We find that genuine 3D understanding does not strictly require 3D-specific training (See \S\ref{sec:scanqa_results}). Furthermore, fixed toolsets are insufficient for open-ended 3D queries, making expressive meta-tools essential. We also observe that a rich structured scene memory often removes the need for direct visual input during spatial reasoning. Lastly, we see that spatial reasoning capabilities of the agent improves naturally with improvements to the underlying foundation models.

More broadly, the goal of training implicit, 3D-native models primarily is to achieve broad generalization and handle long-tail, open-ended queries~\cite{hong20233d_3dllm, wang2023chat3d, cheng2024spatialrgpt, huang2024leo}. However, our findings demonstrate that an explicit, zero-shot framework can successfully satisfy both of these objectives\footnote{Our website provides qualitative examples: \website.}. We posit that future progress in 3D scene understanding may benefit from developing richer explicit scene memories and stronger compositional abstractions for zero-shot reasoning, potentially in combination with advances in 3D-native architectures, rather than relying solely on scaling such architectures.

\section{Related Work}

\begin{figure}[t!]
    \centering
    \includegraphics[width=\linewidth]{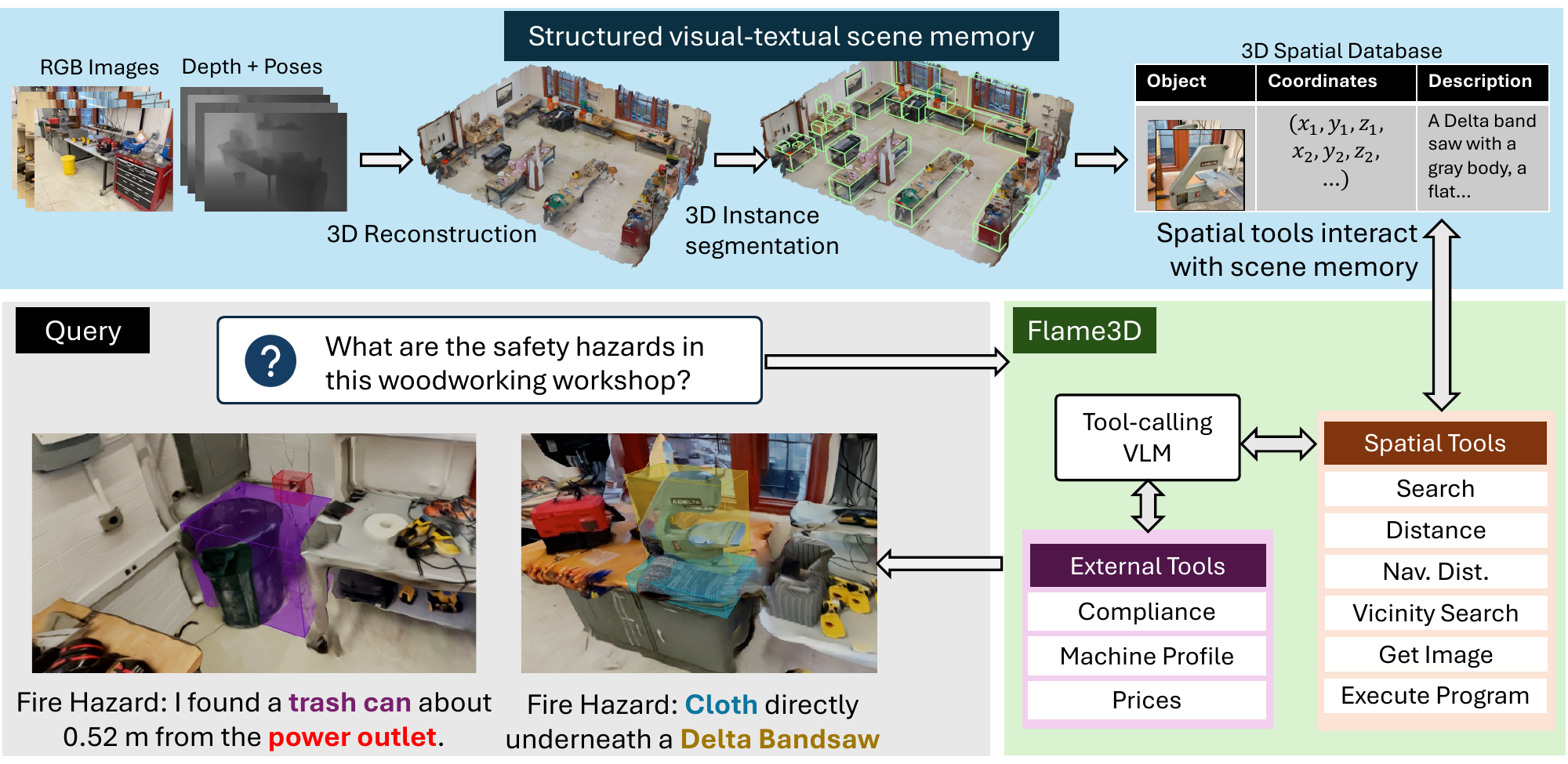}
    \caption{\textbf{Overview of \projectnamecolor{}.} When a natural-language query is received, an off-the-shelf tool-calling vision--language model breaks it down into a sequence of spatial and external tool-calls to produce a grounded answer. The agent composes these inferences by interacting with the \emph{structured scene memory} through a collection of \emph{spatial tools} (search, distance, vicinity search, navigation distance, image retrieval, and code execution) and optional \emph{external tools} (e.g., compliance databases, machine profiles, pricing catalogs, internet search). The scene memory is constructed from posed RGB-D frames as a 3D spatial database. In this example, the agent demonstrates compositional reasoning by identifying fire hazards in a woodworking workshop to localize a trash can near a power outlet and a cloth underneath a bandsaw.}
    \label{fig:method}
\end{figure}

\paragraph{Zero-shot 3D reasoning with off-the-shelf models.}

Most directly related to our work are zero-shot approaches that prompt off-the-shelf MLLMs. One line presents the scene as 2D views, selecting keyframes~\citep{taguchi2025spatialprompting}, rendering bird's-eye maps~\citep{zhang2024agent3dzero}, or allowing active camera manipulation~\citep{zhang2026think3d,li2025see}. A second line exposes structured operations, orchestrating hand-coded geometric operators~\citep{zantout2025sort3d} or using a scene graph for task planning~\citep{rana2023sayplan}. On the evaluation side, ScanQA~\citep{azuma2022scanqa} and SQA3D~\citep{ma2023sqa3d} are the standard 3D QA benchmarks; \benchmarkname{} complements them with complex multi-hop reasoning. Each of these prior systems is focussed on a specific application, fixes its representation and operating set up front. \projectname{} instead retains a structured visual-textual representation of the whole scene and adds meta-abstractions that let the agent synthesize new operations as code at inference time.

\paragraph{Trained 3D scene representations.}
Finetuned 3D large multimodal models~\citep{hong20233d_3dllm,wang2023chat3d,xu2024pointllm,chen2024spatialvlm,chen2024ll3da}
treat 3D as a new modality, tokenizing the point cloud and aligning it into the latent space of an LLM. As discussed in \S\ref{sec:intro}, this approach has issues with obtaining paired 3D-language data, limited compositional generalization, and
catastrophic forgetting risks. Recent work has also questioned the value of point-cloud tokens
for LLM spatial reasoning~\citep{zhang2025point,ma20263real3dqa}. Visual feature fields take a different route, projecting
pretrained encoders into 3D~\citep{jatavallabhula2023conceptfusion,kerr2023lerf,peng2023openscene,werby2024hierarchical,linok2025beyond,gu2024conceptgraphs},
where the underlying object semantics inherit construction-time encoders. \projectname{} sidesteps both approaches by relying
on existing modalities of an off-the-shelf MLLM and exposing the scene as a structured visual-textual representation via
tool calling, so attributes can be edited and operations can be composed during inference.

\paragraph{Agentic LLM with scene memory.}
Our use of tools and meta-tools with scene memory draws on a thread of LLM agents that interleave reasoning with tool calls and synthesize executable code as a way 
of acting in the world~\citep{yao2023react,schick2023toolformer,wang2023voyager,liang2023code,singh2023progprompt,fu2026cap}. The same pattern of language-as-memory 
shows up in embodied AI, both for navigation and long-horizon planning~\citep{ahn2022can,chang2024goat,gervet2023navigating,deitke2022procthor,huang2022innermonologue,park2023generative}. 
\projectname{} applies this perspective to compositional 3D spatial reasoning over a structured 3D scene representation, with tools that the agent 
extends at inference time by generating code.

\section{Method}
\label{sec:method}

\projectname{} takes as input a set of posed RGB-D frames of an indoor scene and a natural-language query, and produces a grounded answer by reasoning over a structured visual-textual scene memory through a collection of spatial and visual abstractions.  The pipeline has three stages (Figure~\ref{fig:method}): (i) constructing a \emph{structured visual-textual scene memory} from the raw scan (\S\ref{sec:scene-memory}), (ii) exposing this memory to a tool-calling MLLM through a set of \emph{spatial and visual abstractions} (\S\ref{sec:abstractions}), and (iii) an \emph{agentic reasoning loop} in which the MLLM composes multi-hop inferences, optionally extending the abstraction set at inference time via meta-abstractions (\S\ref{sec:agent-loop}).

\subsection{Structured Visual-Textual Scene Memory}
\label{sec:scene-memory}

The first stage converts a raw RGB-D scan into a persistent, queryable representation of the scene that an off-the-shelf MLLM can operate on through text and tool calls, without requiring any 3D-specific fine-tuning. Given a set of posed RGB frames and their corresponding depth maps, we first get a 3D reconstruction of the scene.

\paragraph{3D instance segmentation.}
We first build an object inventory by prompting a vision--language model to enumerate objects in each frame and clustering synonymous labels. We then run SAM3~\citep{carion2026sam3segmentconcepts} to extract temporally consistent 2D instance masks, which we then project into 3D using the reconstructed geometry. Finally, we construct a semantic and spatial connectivity graph to merge partial views into persistent 3D \emph{components}. See Appendix~\ref{sec:appendix-instance-seg} for details.

\paragraph{3D spatial database.}
For each component we compute and store: (a)~a 3D bounding box and centroid coordinates, (b)~representative image crops selected from the viewpoints with the best visibility, and (c)~a textual description generated by captioning the selected crops with an off-the-shelf vision--language model.  The caption captures object identity, visual attributes (color, material, shape), and local spatial context.
The components and their attributes are organized into a structured database that we call the \emph{scene memory}.  Each row records the component's unique identifier, its 3D coordinates, its textual description, and pointers to its image crops.  Because the memory is textual, arbitrary non-visual attributes (e.g., price, compliance status, machine specifications) can be appended to any component at any time without retraining or re-encoding the scene, a property shared by neither feature fields nor finetuned 3D-LMMs. The scene memory is stored as a PostGIS spatial database~\cite{postgis} indexed on both 3D coordinates and textual annotations enabling fast spatial and text-based retrieval.

\subsection{Spatial and Visual Abstractions}
\label{sec:abstractions}

Rather than passing the entire scene memory as a monolithic context, \projectname{} exposes it to the MLLM through a small set of composable tool-call abstractions.  Each abstraction encapsulates a specific spatial or visual operation and returns a structured result that the agent can feed into subsequent reasoning steps.

\paragraph{Spatial tools.}
\begin{figure}[!t]
  \centering
  \includegraphics[width=\linewidth]{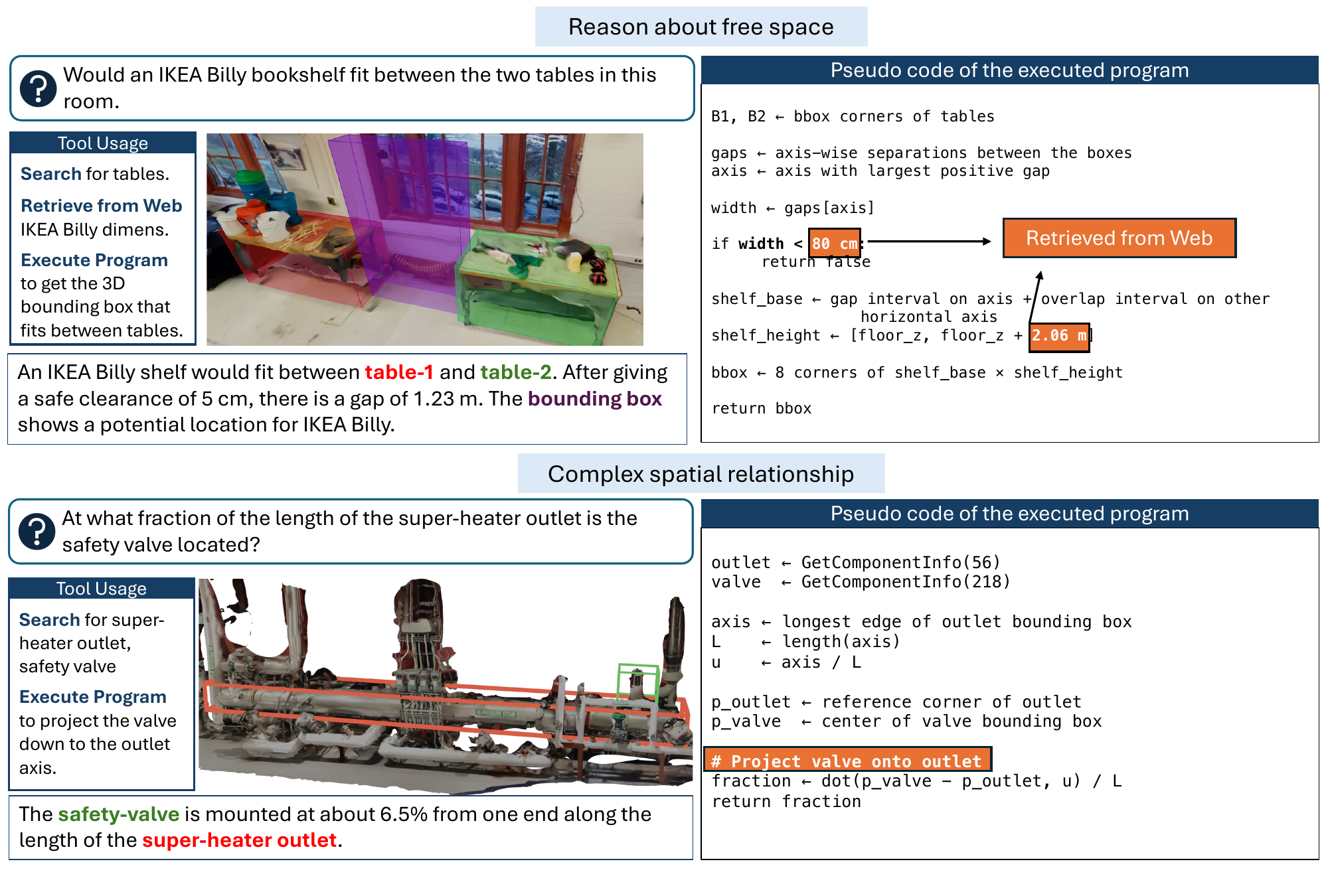}
  \vspace{-2em}    
  \caption{\textbf{Meta-abstraction Tool Use.} Examples of the \emph{Execute program} abstraction. \projectname{} can synthesize tailored Python code at inference time with attributes from external sources, draw arbitrary bounding boxes (top), or compute precise geometric relationships such as projections (bottom).}
  \vspace{-1em}    
  \label{fig:code_examples}
\end{figure}
We provide the following spatial abstractions that operate over the 3D coordinates and structure of the scene memory:  \emph{Search} retrieves components whose textual descriptions match a natural-language query, using BM25~\cite{robertson2009probabilistic} over the component captions.  \emph{Distance} computes the Euclidean distance between the 3D centroids of two specified components. \emph{Navigation distance} computes the traversable path length between two components on the reconstructed floor plane, accounting for obstacles.  \emph{Vicinity search} returns all components within a specified radius of a given component, enabling local spatial reasoning.  \emph{Get image} retrieves the representative image crops for a specified component, allowing the MLLM to perform visual inspection.
Finally, \emph{Execute program} is a meta-tool that lets the agent write and execute arbitrary code over the scene memory at inference time, extending the abstraction set beyond the pre-built tools when the query demands novel spatial or logical operations. As illustrated in Figure~\ref{fig:code_examples}, this allows the agent to construct tailored programs to evaluate complex object insertions (e.g., verifying if a shelf fits into a specific gap) or to compute precise geometric relations (e.g., projecting coordinates to find fractional distances along an axis) directly on the fly. 

\paragraph{External tools.}
For queries that require knowledge beyond the scene itself, \projectname{} can interface with external data sources through additional tool calls.  These include compliance databases (e.g., fire-safety regulations), machine profiles (e.g., equipment specifications), and pricing catalogs.  Because external attributes are integrated as text into the same scene memory, the agent reasons over them with the same abstractions it uses for spatial and visual information, without architectural changes.

\subsection{Agentic Reasoning}
\label{sec:agent-loop}

Given a natural-language query and access to the scene memory through the abstractions defined above, \projectname{} uses a tool-calling MLLM to perform multi-hop compositional reasoning in a ReAct-style loop~\citep{yao2023react}. The agent receives the query together with a system prompt that describes the available tools, their signatures, and the structure of the scene memory.  The prompt encourages the model to plan a sequence of tool calls before acting.

\begin{wrapfigure}{r}{0.4\textwidth}
    \footnotesize
    \texttt{There is a <component\_49>fire extinguisher</component\_49> and a <component\_93>first aid kit</component\_93> in the room...}
    \caption{Example of grounded output with component identifiers.}
    \label{fig:grounded_example}
    \vspace{-1em}
\end{wrapfigure}

\paragraph{Iterative tool use.}
At each step, the MLLM produces either a tool call or a final answer.  Tool calls retrieve information from the scene memory (e.g., searching for relevant components, computing distances, inspecting images, etc.) or from external sources.  The result of each tool call is appended to the agent's context, enabling multi-hop chains. For example, the agent may first search for all power outlets, then query the vicinity of each outlet for combustible materials, then compute distances to produce a ranked list of fire hazards.

\paragraph{Grounded output.}
The agent's final answer references specific components by their identifiers, grounding the response in the 3D scene. This explicit tagging, as illustrated in Figure~\ref{fig:grounded_example}, enables downstream visualization where each referenced object cites specific coordinates of the 3D scene and is linked to its corresponding description and selected image crops.

\section{Experiments}
\label{sec:experiments}

\subsection{Datasets}
\label{subsec:expSetting}

\begin{figure}[!t]
    \centering
    \includegraphics[width=1\linewidth]{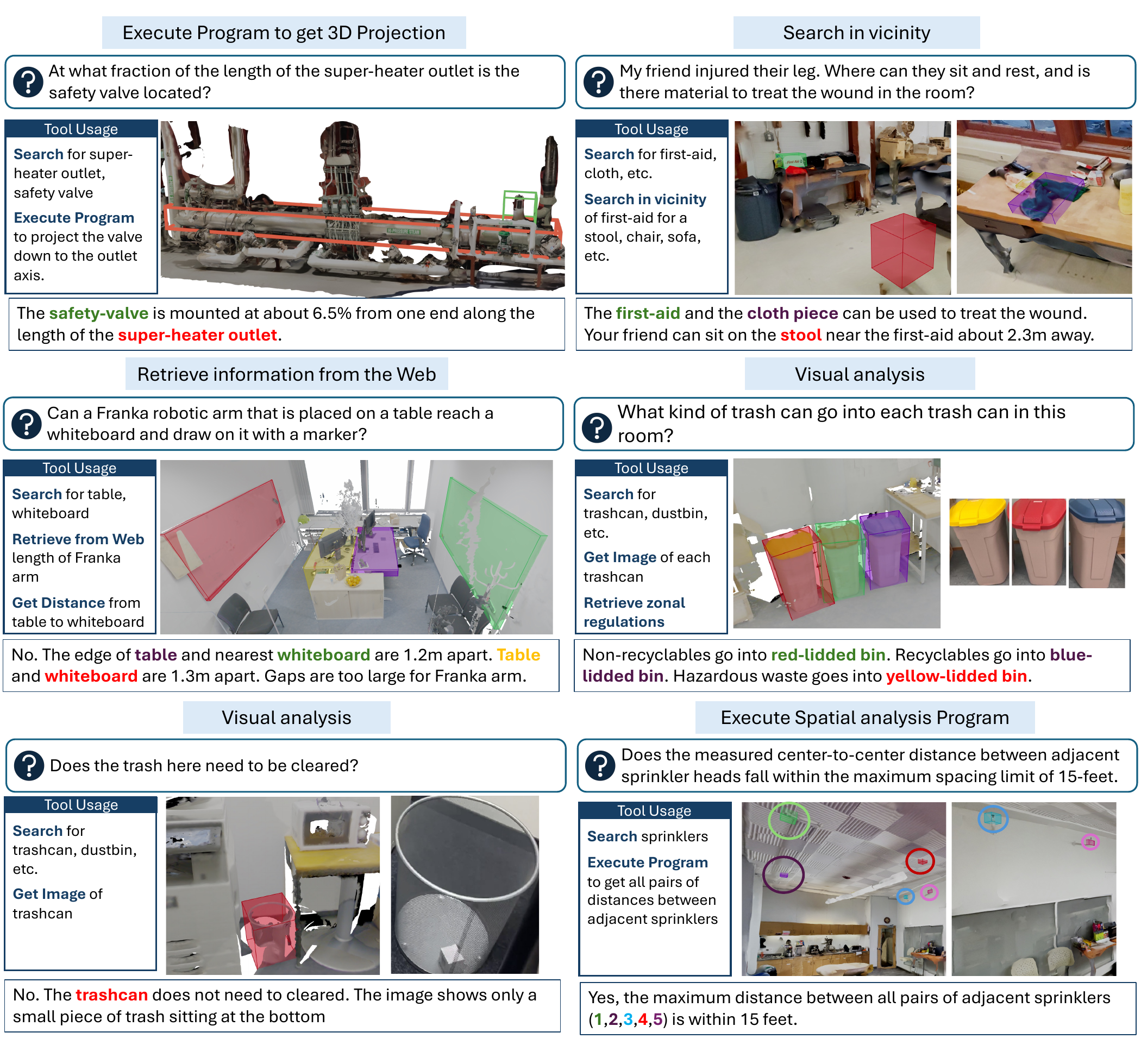}
    \vspace{-2em}    
    \caption{Qualitative examples illustrating \projectname{} answering complex 3D queries by chaining spatial tools, visual inspection, and external knowledge retrieval.}
    \label{fig:qualitativeExample}
    \vspace{-1em}    
\end{figure}

\begin{table}[t]
\centering
\renewcommand{\arraystretch}{1.12}
\resizebox{0.7\columnwidth}{!}{%
\begin{tabular}{lccccc}
\toprule
\textbf{Method}
& \textbf{CIDEr}$\uparrow$
& \textbf{BLEU-4}$\uparrow$
& \textbf{METEOR}$\uparrow$
& \textbf{ROUGE}$\uparrow$
& \textbf{EM}$\uparrow$ \\
\midrule

\multicolumn{6}{l}{\textit{\textbf{Expert models}}} \\
\midrule
ScanQA~\citep{azuma2022scanqa}
& 64.9 & 10.1 & 13.1 & 33.3 & 21.1 \\
3D-VisTA~\citep{zhu20233dvista}
& 69.6 & 10.4 & 13.9 & 35.7 & 22.4 \\

\midrule
\multicolumn{6}{l}{\textit{\textbf{3D LMMs}}} \\
\midrule
3D-LLM~\citep{hong20233d_3dllm}
& 69.4 & 12.0 & 14.5 & 35.7 & 20.5 \\
LEO~\citep{huang2024leo}
& 101.4 & 13.2 & 20.0 & 49.2 & 24.5 \\
Video-3D-LLM~\citep{zheng2025video3dllm}
& 102.1 & 16.2 & 19.8 & 49.0 & 30.1 \\
CVP~\citep{Chen_2026_WACV}
& \textbf{107.1} & 17.8 & 20.8 & \textbf{50.9} & \textbf{31.2} \\

\midrule
\multicolumn{6}{l}{\textit{\textbf{Zero-shot methods}}} \\
\midrule
VideoChat2~\citep{li2024mvbench}
& 49.2 \posdiff{29.24}
& 9.6 \posdiff{24.45}
& 9.5 \posdiff{24.54}
& 28.2 \posdiff{12.27}
& 19.2 \posdiff{7.24} \\

Qwen2.5-VL-7B~\citep{bai2025qwen25vl}
& 53.9 \posdiff{24.54}
& 3.0 \posdiff{31.05}
& 11.4 \posdiff{22.64}
& 29.3 \posdiff{11.17}
& -- \\

SpatialPrompting~\citep{taguchi2025spatialprompting}
& 87.69 \negdiff{-9.25}
& --
& 16.85 \posdiff{17.19}
& 43.39 \negdiff{-2.92}
& 27.34 \negdiff{-0.90} \\

LLaVA-Video~\citep{zhang2024llavavideo}
& 88.7 \negdiff{-10.26}
& 3.1 \posdiff{30.95}
& 17.7 \posdiff{16.34}
& 44.6 \negdiff{-4.13}
& -- \\

\midrule
\textbf{Ours}
& 78.44
& \textbf{34.05}
& \textbf{34.04}
& 40.47
& 26.44 \\

\bottomrule
\end{tabular}}
\caption{\textbf{Performance on the ScanQA validation set.} The smaller text next to the numbers reports the difference between \projectname{} and the corresponding baseline (\projectname{} - baseline). See Appendix~\ref{sec:appendix_scanqa_comparison} for the comprehensive evaluation table.}
\label{tab:scanQAVal}
\vspace{-2em}
\end{table}

\textbf{ScanQA}~\citep{azuma2022scanqa} is a 3D question answering benchmark built on top of real-world indoor RGB-D scans from ScanNet~\cite{dai2017scannet}. Given a reconstructed 3D scene and a natural-language question, the task requires the model to produce an answer grounded in the spatial and semantic content of the scene. The benchmark contains diverse questions about object attributes, locations, counts, and relationships; each question is associated with relevant 3D object annotations when available.

\begin{wrapfigure}{r}{0.3\textwidth}
  \vspace{-2em}
  \begin{center}
    \includegraphics[width=0.3\textwidth]{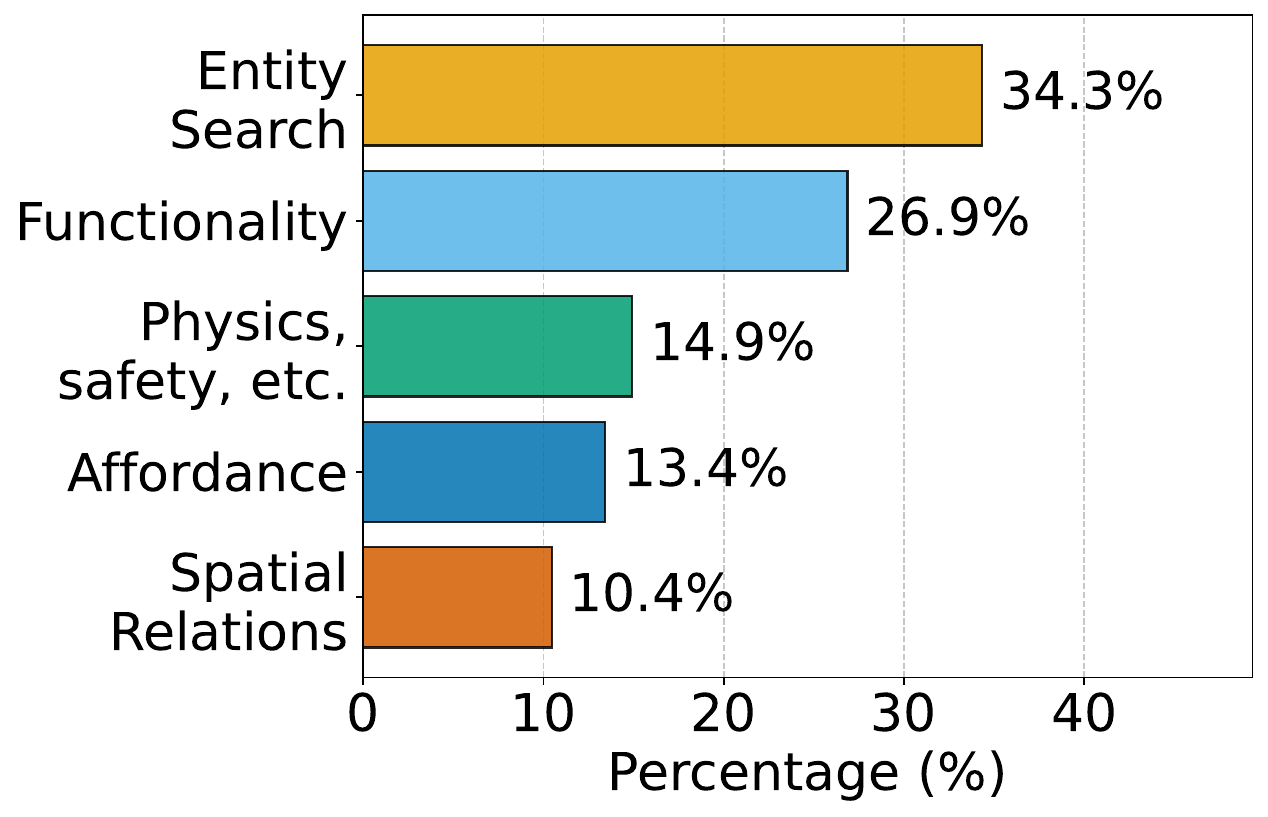}
  \end{center}
  \caption{Distribution of different kinds of questions in \benchmarkname{}.}
  \vspace{-1em}  
\end{wrapfigure}

\textbf{\benchmarknamecolor{}}: 
Although ScanQA provides a valuable foundation for 3D question answering, it lacks complex multi-hop reasoning questions that thoroughly test an agent's spatial understanding. Concurrent studies, such as Real3D-QA~\citep{ma20263real3dqa}, have demonstrated that many questions in existing benchmarks like ScanQA can be answered ``blind'', i.e., without any access to 3D data or visual inputs, by simply relying on language priors. For example, questions in ScanQA like ``What material is the refrigerator made of?'' (answer: ``stainless steel'') or ``What is in the corner of the bath?'' (answer: ``shower'') can often be guessed using common sense rather than genuine 3D spatial reasoning. Furthermore, some questions are ambiguous or ill-posed, such as ``On what part of the cabinet is the beige cabinet placed?'' (answer: ``right kitchen wall'').

To address these limitations, we create \benchmarkname{}, a curated benchmark specifically designed to evaluate complex, multi-hop 3D reasoning. We author these questions on ScanNet++~\citep{yeshwanth2023scannet++} validation scenes. Specifically, we convert the ScanNet++ scenes into spatial scene memory described in Section~\ref{sec:scene-memory}. To facilitate authoring questions for the benchmark, we set up a custom interactive tool and provide the means for each expected answer to be strictly grounded in 3D space (by tagging entities as in Figure~\ref{fig:grounded_example}). Following the taxonomy introduced by recent work~ \citep{sahoo2026conversationalimagesegmentationgrounding}, we categorize our questions into five distinct types: \textit{Entity Search} (identifying and locating specific objects based on open-ended descriptions), \textit{Spatial Relations} (reasoning about geometric arrangements and distances between entities), \textit{Affordance} (reasoning about potential actions an agent can perform within a scene, e.g., ``Where can I sit?''), \textit{Functionality} (reasoning about the utility of objects, e.g., ``Find me something to clean the room''), and \textit{Physics, Safety, etc.} (requiring reasoning about physical properties and safety constraints). 
For more details please refer to Appendix~\ref{sec:benchmark-collection}.




\subsection{Qualitative Examples}

Figure~\ref{fig:qualitativeExample} presents several qualitative examples demonstrating the versatility of \projectname{} in answering diverse spatial, visual, and multi-hop queries. By actively selecting and chaining tools, the agent tackles scenarios that would be difficult for models lacking explicit 3D abstractions or external knowledge access. For instance, to answer precise geometric queries such as determining the fractional position of a safety valve along a superheater outlet, or verifying if sprinkler spacing meets a 15-foot safety code, the agent uses the \emph{Execute program} tool to perform custom mathematical projections and pair-wise distance checks directly on the scene's 3D coordinates. In scenarios requiring context about human actions, such as finding a resting spot near first-aid supplies, the agent chains \emph{Search} and \emph{Vicinity search} to identify nearby seating options. Furthermore, \projectname{} seamlessly integrates external knowledge and visual verification into its reasoning loop. When asked if a Franka robotic arm placed on a table can reach a whiteboard, it queries the web for the robot's arm length and compares it against the computed 3D distance between the objects. Similarly, for compliance and maintenance queries, such as determining waste sorting rules or checking if a trash bin needs clearing, the agent retrieves relevant regulations and uses the \emph{Get image} tool to visually inspect the state of the bins. These examples highlight how the composition of spatial, visual, and external tools enables robust, grounded 3D reasoning.

\subsection{Results and Insights}
\label{sec:scanqa_results}

For ScanQA, we evaluate answer quality using standard metrics: exact match (EM), BLEU-4~\citep{papineni2002bleu}, METEOR~\citep{banerjee2005meteor}, ROUGE-L~\citep{lin2004rouge}, and CIDEr~\citep{vedantam2015cider}. For \benchmarkname{}, we evaluate the models on two primary axes: Response Text Evaluation (AI-Judge, METEOR) and Object Grounding Evaluation (F1 Score, Precision, Recall of predicted component IDs); see Appendix~\ref{sec:benchmark-metrics} for the rationale behind these choices. Through our experiments, we identify several key insights regarding agentic 3D spatial reasoning.

\begin{wrapfigure}{r}{0.5\textwidth}
  \vspace{-1.5em}
  \centering
  \includegraphics[width=\linewidth]{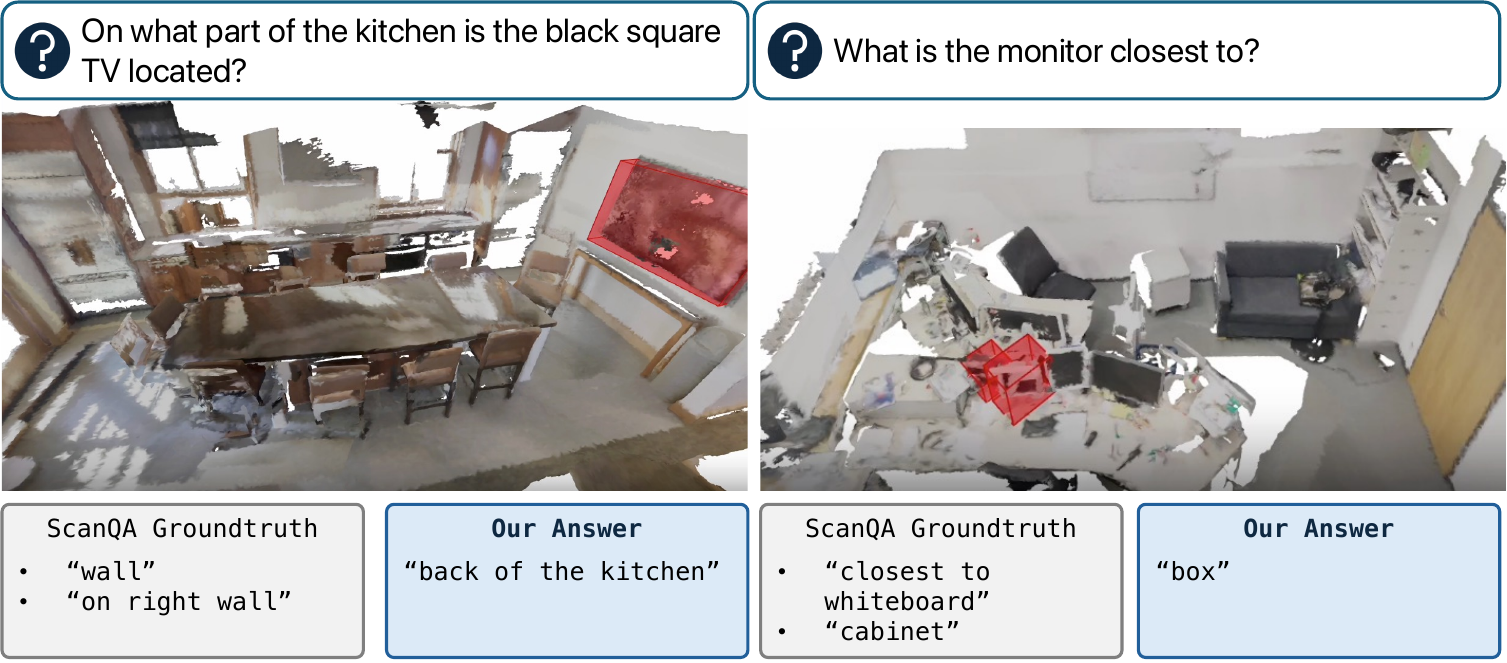}
  \caption{Our method often generates contextually valid answers but are penalized by ambiguity in ScanQA ground-truth.}
  \label{fig:scanqa_qual}
  \vspace{-1em}
\end{wrapfigure}

\paragraph{Genuine 3D understanding may not require training on 3D--language data.}
Table~\ref{tab:scanQAVal} compares \projectname{} with expert 3D QA models, 3D LMMs, and zero-shot 2D LMM-based methods on the ScanQA validation set. Our method is competitive with and often beats methods trained or fine-tuned specifically on ScanNet and ScanQA datasets (e.g., CVP~\citep{Chen_2026_WACV} includes ScanQA in its training set). Our BLEU-4 and METEOR scores are significantly higher than other methods. This demonstrates that zero-shot methods can perform on par with trained models, suggesting that genuine 3D spatial understanding might not strictly require training on 3D--language data. Figure~\ref{fig:scanqa_qual} shows that \projectname{} generates reasonable answers for queries in ScanQA, but these are penalized due to ambiguity in the ScanQA ground truth\footnote{We provide more such examples in Supplementary Material}. This indicates that the true capacity of the system for 3D understanding is even stronger than quantitative metrics suggest. Furthermore, qualitative examples from Figure~\ref{fig:qualitativeExample} clearly demonstrate that \projectname{} can answer even complex and long-tail spatial queries. 

\begin{figure*}[t]
    \centering
    \begin{tabular}{cc @{\hspace{1cm}} ccc}
        \multicolumn{2}{c}{Response Text Evaluation} & \multicolumn{3}{c}{Object Grounding Evaluation} \\
        \cmidrule(r){1-2} \cmidrule(l){3-5}
        
        \begin{subfigure}[b]{0.16\textwidth}
            \centering
            \includegraphics[width=\textwidth]{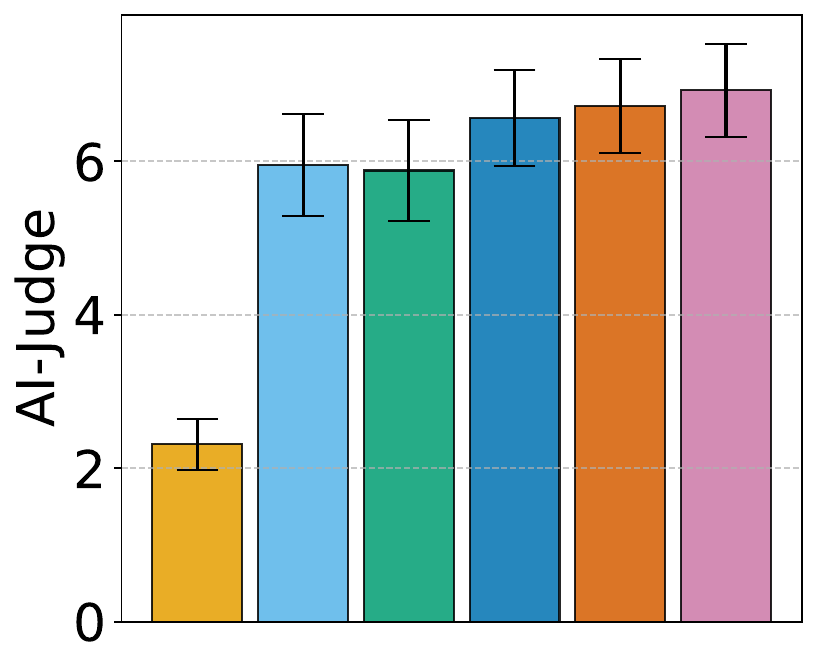}
            \caption{AI-Judge}
        \end{subfigure} &
        \begin{subfigure}[b]{0.16\textwidth}
            \centering
            \includegraphics[width=\textwidth]{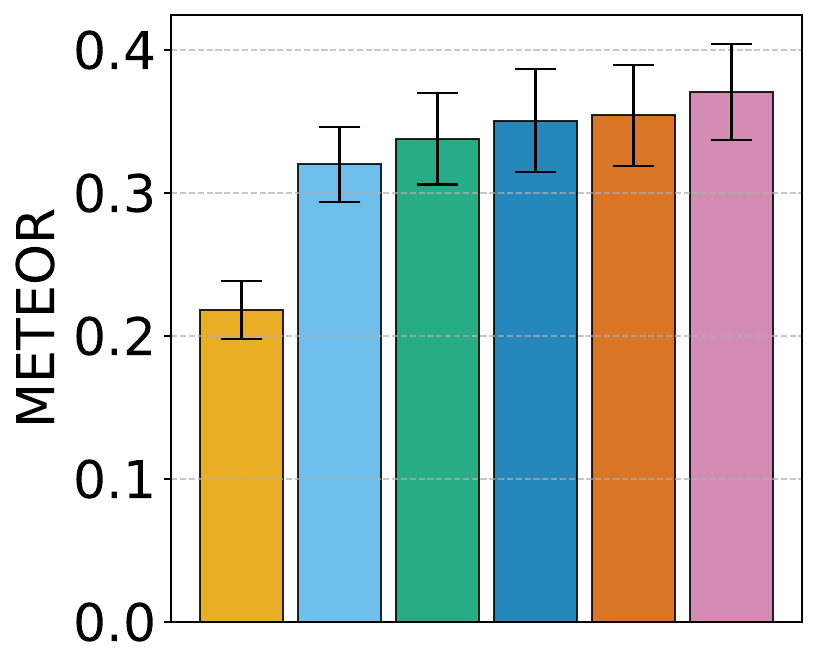}
            \caption{METEOR}
        \end{subfigure} & 
        
        \begin{subfigure}[b]{0.16\textwidth}
            \centering
            \includegraphics[width=\textwidth]{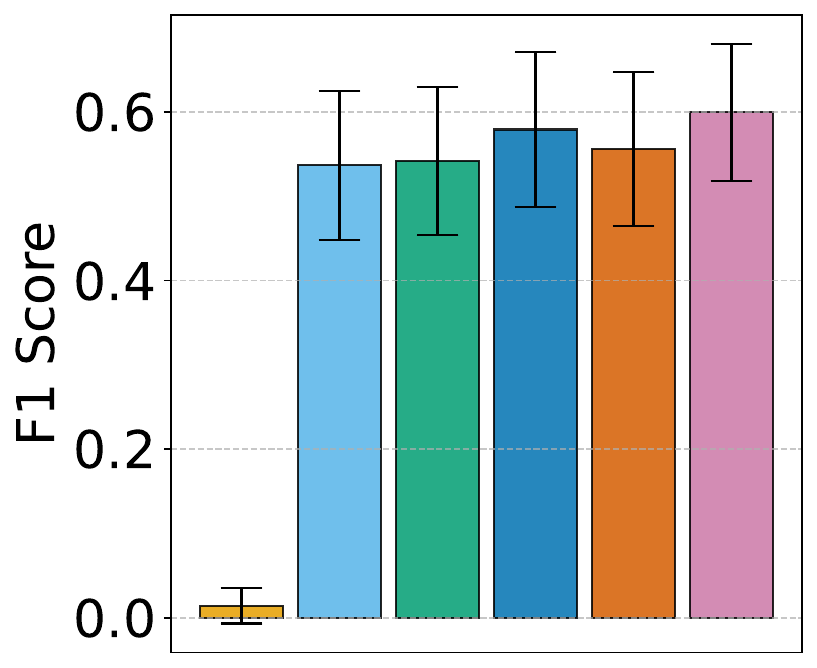}
            \caption{F1 Score}
        \end{subfigure} &
        \begin{subfigure}[b]{0.16\textwidth}
            \centering
            \includegraphics[width=\textwidth]{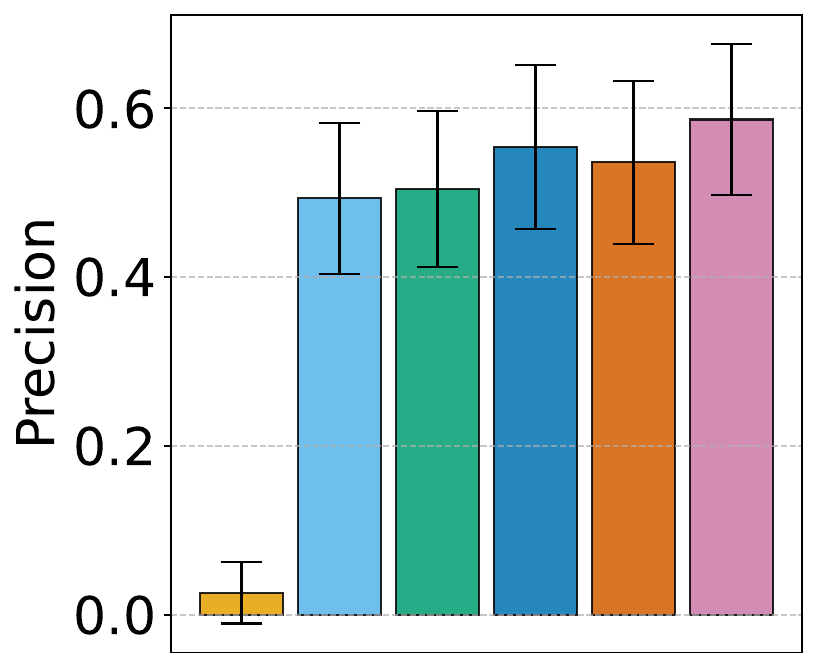}
            \caption{Precision}
        \end{subfigure} &
        \begin{subfigure}[b]{0.16\textwidth}
            \centering
            \includegraphics[width=\textwidth]{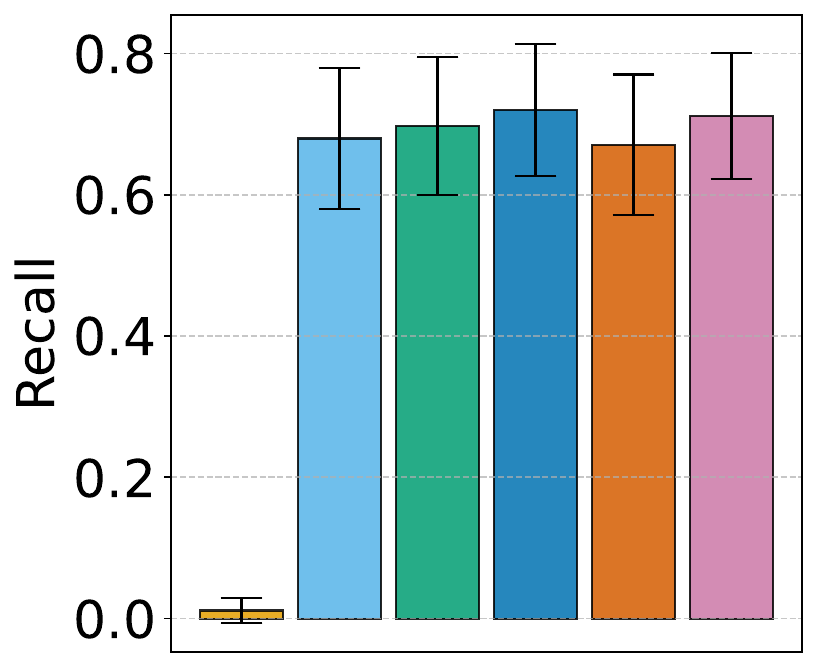}
            \caption{Recall}
        \end{subfigure}
    \end{tabular}

    \includegraphics[height=2.5cm, width=0.85\textwidth, keepaspectratio]{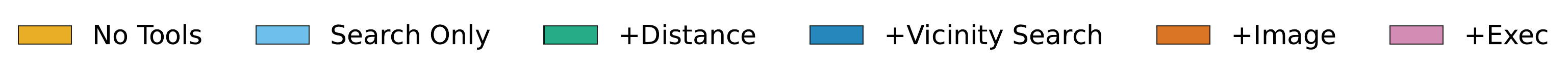}
    \caption{\textbf{Impact of spatial reasoning tools on model performance.} The results illustrate the incremental performance gains achieved by integrating spatial tools.}
    \vspace{-1em}
    \label{fig:evaluation_metrics}
\end{figure*}

\paragraph{Fixed toolsets are insufficient for open-ended 3D queries.}
Figure~\ref{fig:evaluation_metrics} shows the results of our ablation study. The improvement in metrics with the progressive addition of tools conveys that relying on a predefined, fixed set of tools (like simple distance or vicinity searches) is fundamentally insufficient to cover the diverse, long tail of multi-hop 3D queries. To effectively address the full spectrum of spatial reasoning, the agent requires a meta-tool, such as our \emph{Execute program} tool, that allows it to dynamically synthesize custom geometric and logical operations on the fly. 

\paragraph{Visual input is often not required during spatial reasoning.}
Interestingly, direct visual input is often not required to answer the majority of spatial queries. The reason vision can be safely omitted in these cases is that during the offline construction of the scene memory, we already generate detailed captions for the component crops. These captions capture most of the semantic attributes needed for typical queries. However, for long-tail or rare queries, visual access remains important. For example, to answer ``What kind of plants are in the apartment?'', the agent might need to actively retrieve and inspect images if the offline captions lack specific botanical species classifications.

\paragraph{Sole reliance on language priors fails.}
The ``No Tools'' baseline in Figure~\ref{fig:evaluation_metrics} shows a configuration where no spatial tools are made available to the agent and it is asked to answer based on its language priors. We see that this performs exceptionally poorly. This starkly contrasts with performance on traditional benchmarks like ScanQA, where models can often guess or memorize answers using language priors alone~\cite{ma20263real3dqa}. As highlighted by the qualitative examples in Figure~\ref{fig:qualitativeExample} and the method section, resolving queries in \benchmarkname{} requires explicit spatial memory.

\paragraph{Meta tool is surprisingly powerful but not sufficient.}

\begin{figure*}[t]
    \centering
        \begin{tabular}{cc @{\hspace{1cm}} ccc}
            \multicolumn{2}{c}{Response Text Evaluation} & \multicolumn{3}{c}{Object Grounding Evaluation} \\
            \cmidrule(r){1-2} \cmidrule(l){3-5}
            
            \begin{subfigure}[b]{0.16\textwidth}
                \centering
                \includegraphics[width=\textwidth]{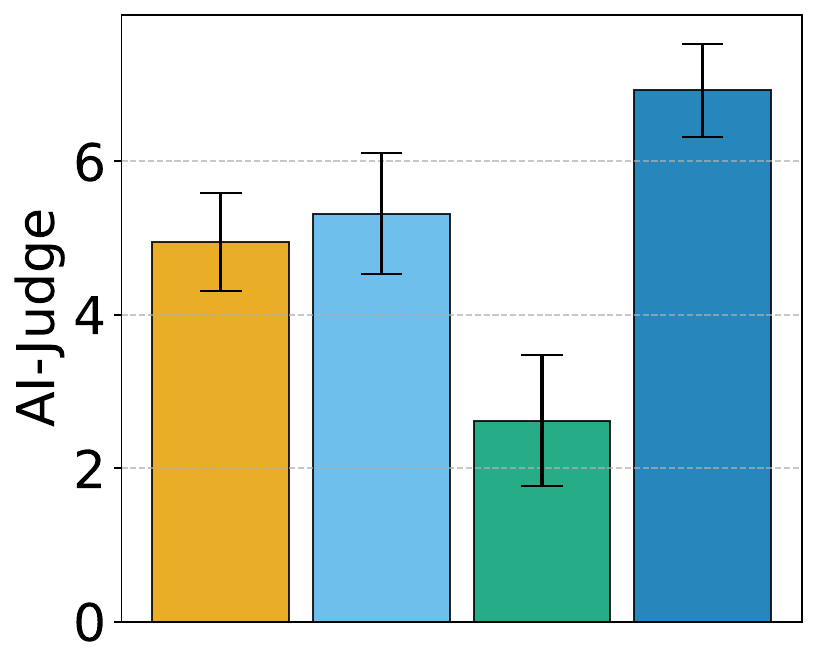}
                \caption{AI-Judge}
            \end{subfigure} &
            \begin{subfigure}[b]{0.16\textwidth}
                \centering
                \includegraphics[width=\textwidth]{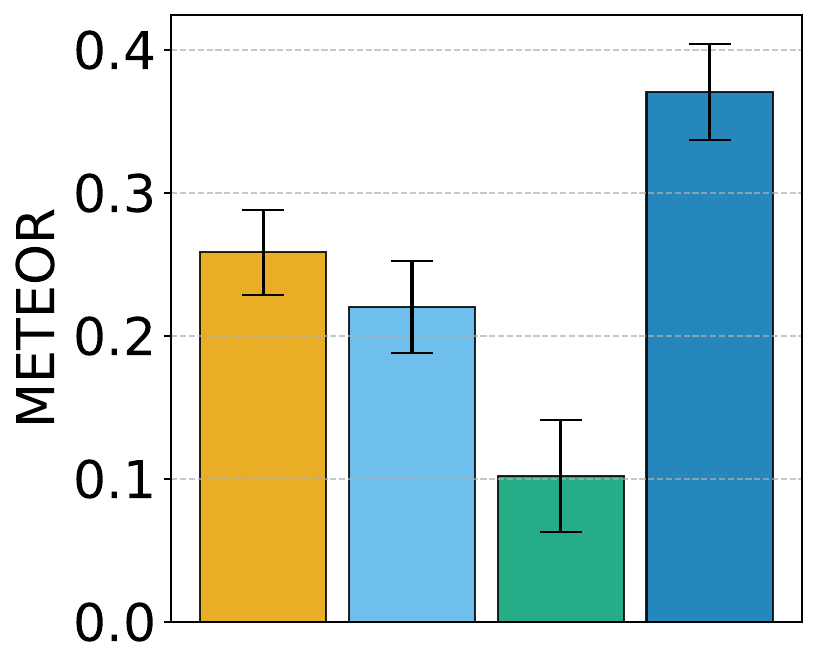}
                \caption{METEOR}
            \end{subfigure} & 
            
            \begin{subfigure}[b]{0.16\textwidth}
                \centering
                \includegraphics[width=\textwidth]{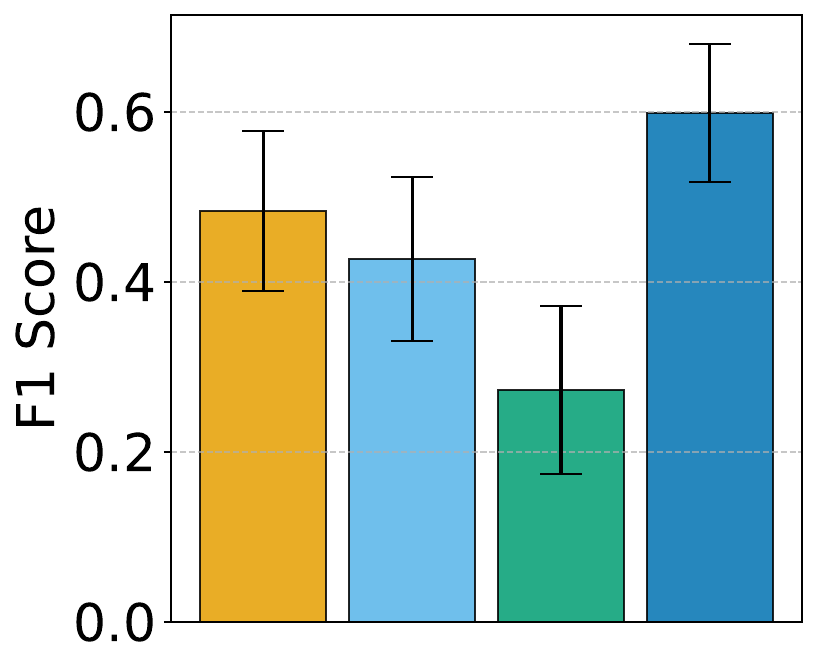}
                \caption{F1 Score}
            \end{subfigure} &
            \begin{subfigure}[b]{0.16\textwidth}
                \centering
                \includegraphics[width=\textwidth]{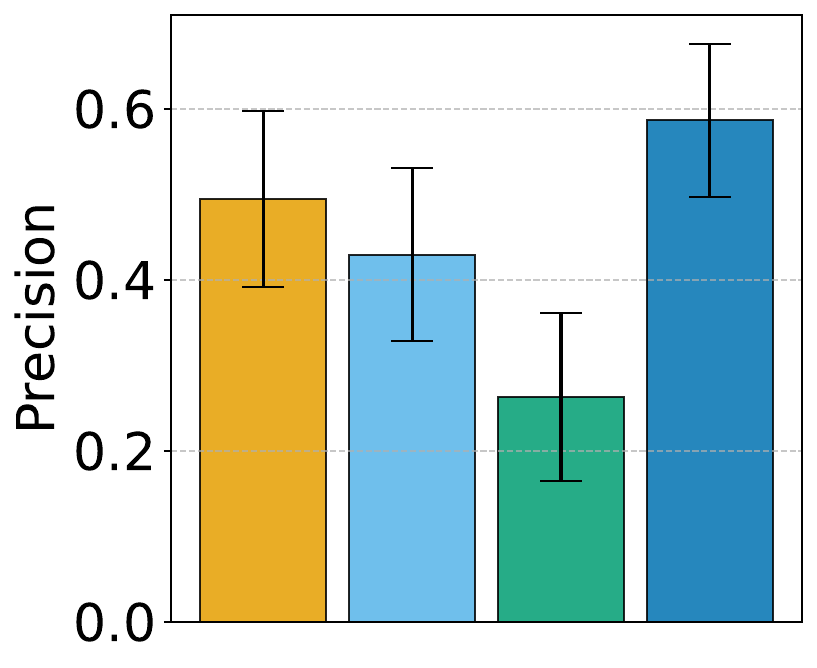}
                \caption{Precision}
            \end{subfigure} &
            \begin{subfigure}[b]{0.16\textwidth}
                \centering
                \includegraphics[width=\textwidth]{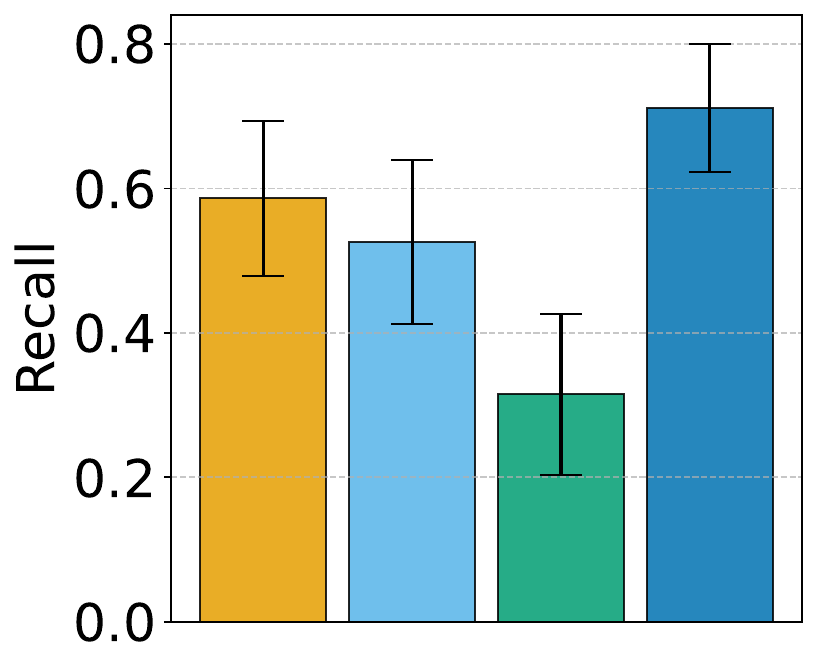}
                \caption{Recall}
            \end{subfigure}
        \end{tabular}%

    \vspace{0.5em} 

    \includegraphics[height=2.5cm, width=0.8\textwidth, keepaspectratio]{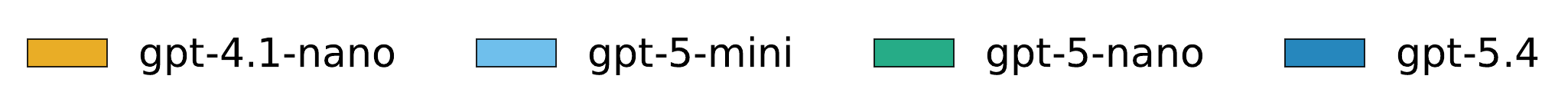}
    
    \vspace{-1em}    
    \caption{\textbf{Comparison across different foundation models.} Modern, larger models consistently demonstrate superior spatial reasoning and object grounding capabilities.}
    \vspace{-2em}    
    \label{fig:model_comparison}
\end{figure*}

\begin{wrapfigure}{r}{0.3\textwidth}
    \centering
    \vspace{-1em}
    \small Response Text Evaluation \\
    \vspace{2pt}
    \begin{subfigure}[b]{0.14\textwidth}
        \centering
        \includegraphics[width=\textwidth]{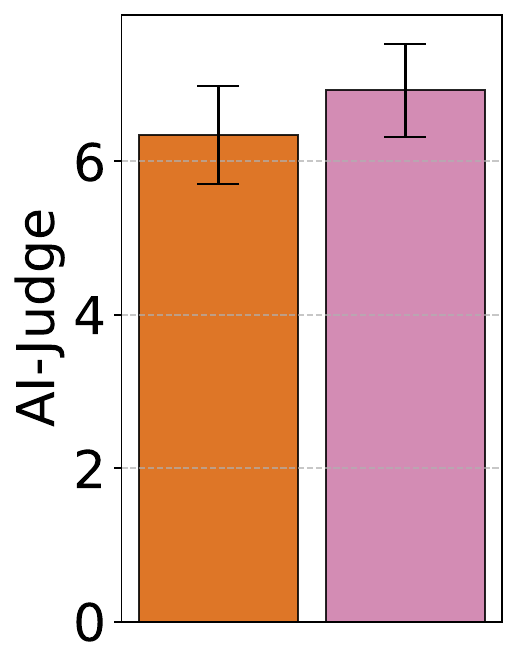}
        \caption{\scriptsize AI-Judge}
    \end{subfigure}%
    \hfill
    \begin{subfigure}[b]{0.14\textwidth}
        \centering
        \includegraphics[width=\textwidth]{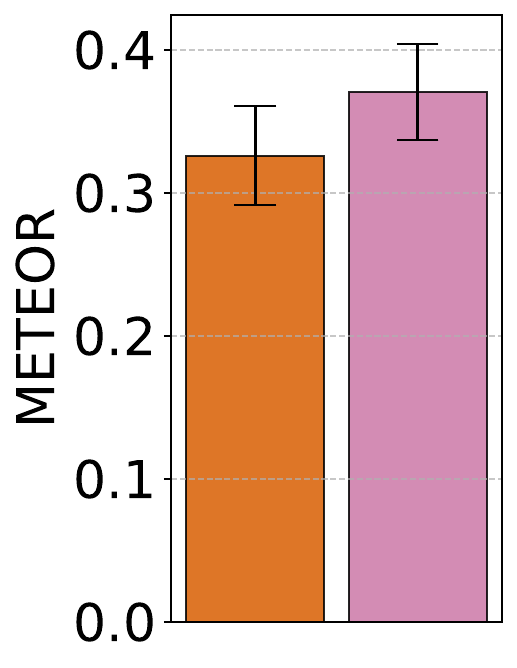}
        \caption{\scriptsize METEOR}
    \end{subfigure}
    
    \small Object Grounding Evaluation \\
    \vspace{2pt}
    \begin{subfigure}[b]{0.14\textwidth}
        \centering
        \includegraphics[width=\textwidth]{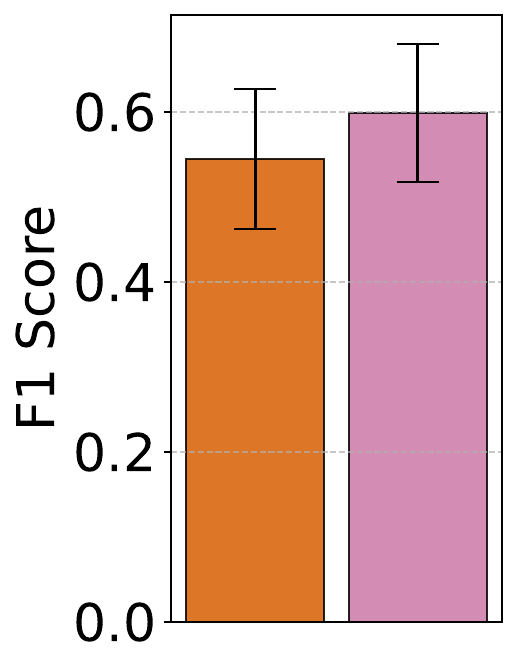}
        \caption{\scriptsize F1 Score}
    \end{subfigure}%
    \hfill
    \begin{subfigure}[b]{0.14\textwidth}
        \centering
        \includegraphics[width=\textwidth]{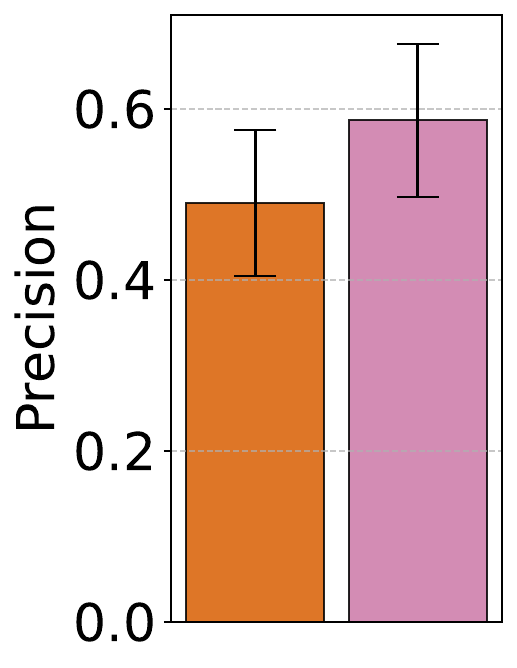}
        \caption{\scriptsize Precision}
    \end{subfigure}

    \includegraphics[width=0.28\textwidth, keepaspectratio]{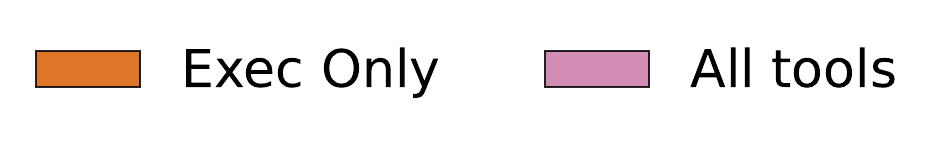}
    
    \caption{Full tool suite provides consistent performance gains over relying solely on the meta tool.}
    \label{fig:meta_tool_ablation}
    \vspace{-4em}
\end{wrapfigure}

We evaluate an ``Execute Only'' baseline where the agent relies entirely on the \emph{Execute program} tool. Note that this tool implicitly has access to all the other spatial tools, except the \emph{Get Image} tool. As shown in Figure~\ref{fig:meta_tool_ablation}, the surprising insight is that this execute-only baseline takes us a long way. Because the meta-tool is highly expressive, it effectively subsumes many geometric tools and establishes a strong foundation for reasoning. However, it does not take us all the way. It cannot, for example, subsume tools like image retrieval. Consequently, providing the full suite of tools yields better results possibly by reducing the cognitive load on the agent and minimizing programming errors during code synthesis. Furthermore, relying purely on arbitrary code execution introduces general privacy and security concerns, making a hybrid toolset more practical for real-world deployment.

\paragraph{Scaling foundation models improves spatial reasoning.}
To understand how the choice of the underlying foundation model impacts spatial reasoning, we evaluate \projectname{} across different language models. As shown in Figure~\ref{fig:model_comparison}, larger models perform significantly better than their smaller or older counterparts. This trend is highly advantageous for our approach: unlike 3D-native models, our system's capabilities naturally scale simply by upgrading to better general-purpose foundation models as they become available.

\section{Discussions}
\label{sec:limitation}

\label{sec:limitations}
\textbf{Limitations.}  The primary limitation of \projectname{} is that the overall performance is dependent on the initial scene memory construction. 
If the underlying models used to detect and segment objects fail to identify some entities during this stage, for example, the query mechanism will be unable 
to locate or reason about those entities. However, we anticipate that the 2D vision-language foundation models and segmentation tools used to construct this 
scene memory will continue to see rapid development.

\label{sec:ethicsAndImpact}
\textbf{Broader Impact.} To scale up \projectname{} across multiple real-world locations, while still maintaining the privacy needs of individual spaces, it could be integrated with a spatial federated system~\cite{openflame_ismar, openflame_nines, openflame_hotos}. For example, \projectname{} could be the spatial intelligence service within a mapserver in OpenFLAME~\cite{openflame_nines}.

\section{Conclusion}

In this work, we introduced \projectnamecolor{}, a training-free, tool-augmented framework for grounded 3D spatial reasoning. By leveraging a structured scene memory, \projectname{} provides a composable and interpretable way of providing compositional 3D reasoning. While achieving competitive results on standard benchmarks, our results and insights drawn from \benchmarkname{} queries highlight the true strength of \projectname{}: the ability to answer very sophisticated, multi-hop queries that require precise geometric 3D reasoning and external knowledge.

Finally, our insights challenge the conventional wisdom that to solve complex, unstructured and long-tailed reasoning over 3D scenes necessitates specialized training approach. By demonstrating that an agentic LLM can tackle spatial reasoning tasks using a structural spatial memory with a well-design toolset, we highlight a promising alternative for advancing spatial 3D reasoning.

\bibliographystyle{unsrtnat}
\bibliography{references}

\clearpage

\appendix

\section{3D Instance Segmentation Pipeline}
\label{sec:appendix-instance-seg}

This section details the multi-stage pipeline that converts posed RGB frames into persistent 3D instance components referenced in \S\ref{sec:scene-memory}.  The pipeline consists of multiple stages, executed sequentially.

\subsection{Object Inventory}

For each RGB frame in the scan, we prompt an off-the-shelf vision--language model (GPT-4o-mini by default) with the instruction \emph{``Identify all tangible objects in this image''}, and parse the response into a list of object labels.  This produces a per-frame inventory of the objects visible in each viewpoint. Note that this stage only identifies objects without segmenting them.

\subsection{Label Normalization}

The raw per-frame labels are noisy. For example, the same physical object may be called ``trash can'', ``garbage bin'', or ``waste basket'' across different frames.  We normalize labels in three steps:

\begin{enumerate}
    \item \textbf{Lemmatization.}  Each label is lowercased and lemmatized as a noun using spaCy, preventing verb-form artifacts (e.g., ``saw'' $\to$ ``saw'' rather than ``see'').
    \item \textbf{CLIP embedding.}  The unique lemmatized labels are embedded with OpenCLIP~\cite{ilharco_gabriel_2021_5143773} to obtain dense semantic representations.
    \item \textbf{Agglomerative clustering.}  We cluster the embeddings using average-linkage agglomerative clustering~\cite{sokal1958statistical} with a cosine-distance threshold (default 0.05).  Within each cluster, the label whose embedding is closest to the cluster centroid is selected as the canonical representative, and all labels in the cluster are replaced by it.
\end{enumerate}

After normalization, we perform \emph{temporal hole filling}: for each canonical label, short gaps (up to 3 consecutive frames) in its frame sequence are backfilled, ensuring that brief occlusions do not fragment the label's temporal span.  Finally, we build an inverted index mapping each canonical label to its contiguous frame subsequences (runs of at least 5 frames), yielding the \emph{objects-to-frames} index that drives the next stage.

\subsection{SAM3 Video Segmentation}

For each (object, frame-subsequence) pair in the objects-to-frames index, we run SAM3~\citep{carion2026sam3segmentconcepts}, a text-prompted video segmentation model.  The object's canonical label is provided as a text prompt on the first frame of the subsequence, and SAM3 propagates segmentation masks across all frames, producing temporally consistent tracking IDs for each detected instance.  The output is a set of per-frame COCO-RLE masks, each annotated with its SAM3 tracking ID (\texttt{obj\_id}), ensuring that the same physical instance receives the same ID across the subsequence.

\subsection{2D-to-3D Mask Association}

A COLMAP~\cite{schoenberger2016sfm} model is constructed from posed RGB-D input images. Each per-frame mask must be grounded in the 3D reconstruction.  For every frame that appears in both the SAM3 output and the COLMAP model, we:

\begin{enumerate}
    \item Construct a \emph{mask canvas} using a painter's algorithm: masks are sorted by bounding-box area (largest first) and painted onto an $(H \times W)$ index map, so that smaller masks on top override larger background masks.
    \item Project all COLMAP 2D feature points (whose 3D correspondences are known) onto the canvas.  Each feature point's pixel location is looked up in the index map to determine which mask (if any) it falls inside.
    \item Accumulate the 3D point IDs for each (object, sequence, instance) triple across all frames in the subsequence.
    \item Apply a per-mask DBSCAN~\cite{dbscan} pass on the accumulated 3D coordinates (default $\varepsilon = 0.5$\,m, $\mathit{min\_samples} = 5$) to discard spatial outlier points caused by noisy 2D projections.
\end{enumerate}

The result is a dictionary mapping each instance triple (object slug, sequence index, SAM3 tracking ID) to its set of associated 3D point IDs.

\subsection{Mask Connectivity Graph}

The same physical entity may appear in multiple object-level subsequences (e.g., once under the label ``chair'' and once under ``seat'') or across disjoint temporal segments.  We merge these duplicate observations by constructing a \emph{mask connectivity graph}:

\paragraph{Nodes.}  Each node corresponds to a unique (sequence, object-slug, SAM3 obj-id) triple with its associated set of 3D point IDs.

\paragraph{Edges.}  We construct a sparse binary incidence matrix $M \in \{0,1\}^{N \times P}$ where $N$ is the number of nodes and $P$ is the number of unique voxel cells after discretizing the point cloud onto a regular grid.  The intersection matrix $M \cdot M^\top$ gives the pairwise overlap counts in a single sparse matrix multiplication.  Two nodes $i, j$ are connected by an edge if their Jaccard similarity exceeds a threshold~$\tau$:
\begin{equation}
    J(i,j) \;=\; \frac{|S_i \cap S_j|}{|S_i \cup S_j|} \;\geq\; \tau
\end{equation}
where $S_i$ and $S_j$ are the voxel-occupancy sets of the two nodes.  Each node's raw point IDs are projected onto a voxel grid (default $50$\,cm side length) spanning the full COLMAP bounding box, so that Jaccard is computed over spatial volume.

\paragraph{CLIP semantic guard.}  For each candidate edge, we extract the best-view masked crop of each node (the frame with the most COLMAP points overlapping the node's 3D set) and compute its OpenCLIP ViT-H-14 image embedding.  Edges whose cosine distance exceeds a threshold (default $0.8$) are rejected, preventing visually dissimilar objects from being merged even if they share 3D points due to reconstruction noise.

\paragraph{Constrained connected components.}  Na\"ive connected-component extraction could transitively merge two nodes that SAM3 explicitly identified as distinct instances within the same sequence.  We therefore use a \emph{constrained union-find}: before merging two components, we check whether their (sequence, object-slug) key-sets overlap; if so, the merge is rejected.  Edges are processed in decreasing order of Jaccard confidence so that the strongest associations are established first.  Each resulting connected component becomes a persistent 3D \emph{component} in the scene memory.

\subsection{Component Cleaning}

A final DBSCAN pass (default $\varepsilon = 0.1$\,m, $\mathit{min\_samples} = 5$) is applied to the 3D coordinates of each connected component.  Points assigned to the noise cluster (label $= -1$) are discarded.  Components with fewer than a minimum number of inlier points (default 20) are removed entirely.  Optionally, components containing multiple distinct DBSCAN clusters can be split into separate components, though by default only noise removal is performed.

\subsection{Component Attributes}

For each surviving component, we compute (a) an axis-aligned 3D bounding box from its inlier points, (b)~select the top-$k$ representative image crops ranked by the number of visible component points in each frame, and (c)~caption the crops with a VLM (using the prompt \emph{``These images show different views of the same object\ldots Provide a concise caption''}).  The component's 3D coordinates, bounding box, image crops, and caption are stored as a row in the \emph{scene memory} database.
\section{\benchmarkname{} Benchmark}
\label{sec:benchmark-collection}

The authors create \benchmarkname{} benchmark through a custom interactive interface that loads a 3D scan into a viewer and lets us construct multi-hop questions whose answers are grounded to specific entities in the scene. The protocol is designed to produce queries that require genuine 3D reasoning rather than linguistic shortcuts~\citep{ma20263real3dqa}, and to bring out the role of necessary spatial and visual abstractions.

\begin{figure}
    \centering
    \fbox{\includegraphics[width=0.85\linewidth]{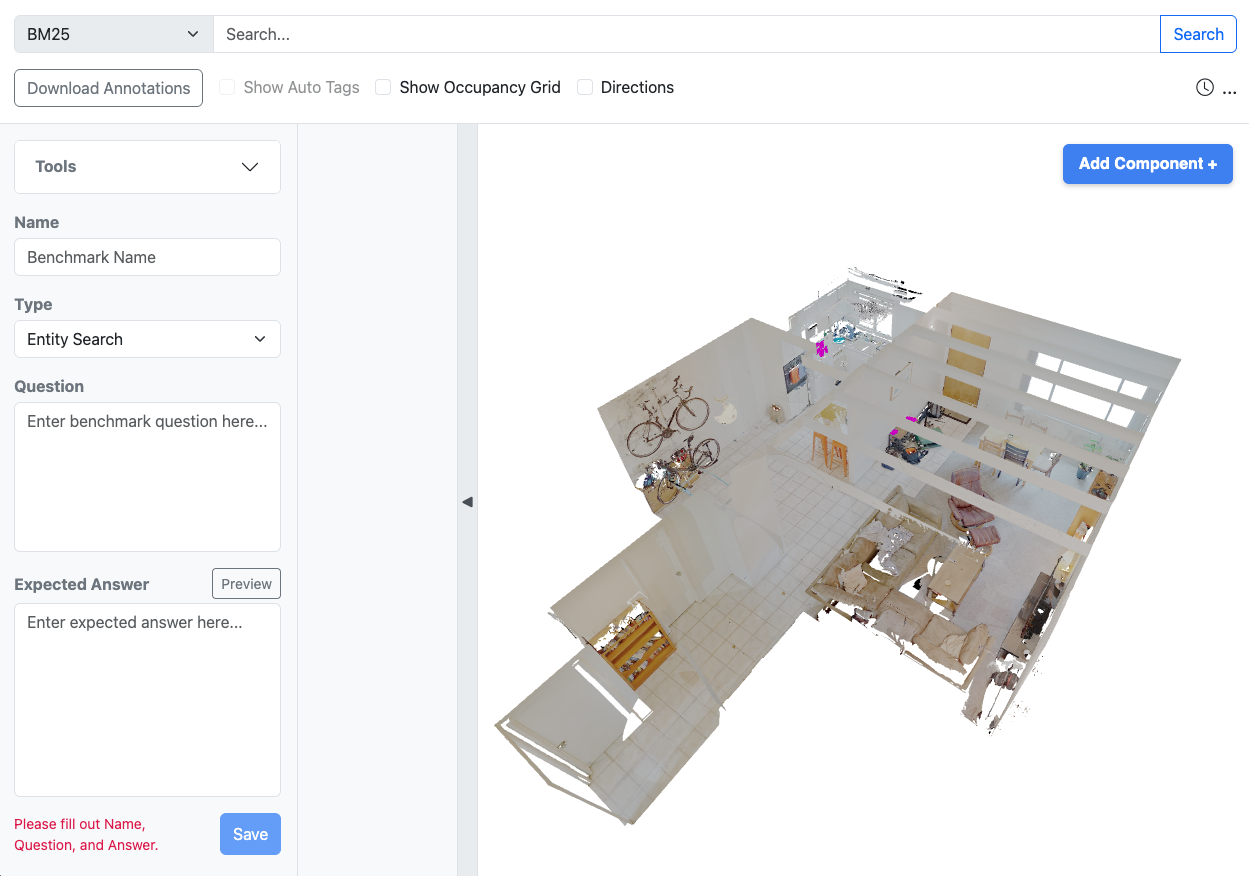}}
    \caption{Interface of the benchmark collection tool.}
    \label{fig:benchmarkCollectionInterface}
\end{figure}

\subsection{Interface}

The collection interface, shown in Figure~\ref{fig:benchmarkCollectionInterface}, presents the scan in a 3D viewer alongside a sidebar that lists annotated bounding boxes, exposes BM25 search over object captions, and provides auxiliary spatial queries. Annotators interact with the scene by:

\begin{itemize}
    \item \textbf{Loading annotations.} A ``Download Annotations'' action fetches all bounding boxes for the dataset and renders them in the viewer; an optional overlay shows auto-generated tags above each box.
    \item \textbf{Searching for components.} A BM25 search over object captions populates the sidebar with matches and focuses the camera on each result.
    \item \textbf{Identifying components.} Clicking a bounding box (in the viewer or the sidebar) opens a Component Details panel showing a cropped image, caption, and physical dimensions, and reveals the \emph{Component ID} required for grounding answers.
\end{itemize}

\subsection{Question and Answer Format}

Each item in \benchmarkname{} consists of a natural-language \textbf{Question} and an \textbf{Expected Answer} that contains the ideal rationale in text. To allow strict grounding, all object references in the answer are wrapped in component tags using the syntax \texttt{<component\_ID>text</component\_ID>}, following conversational grounded supervision. For example:

\begin{quote}
\texttt{You can use the <component\_23>coffee machine</component\_23> located next to the <component\_45>fridge</component\_45>.}
\end{quote}

A preview mode parses these tags into clickable links that snap the camera onto the referenced component, allowing the annotator to verify each grounding before saving. Tags whose IDs do not exist in the scene are flagged inline.

\subsection{Auxiliary Tools Available}

To help us reason about spatial geometry while drafting questions, the interface exposes the same family of spatial abstractions that the agent uses at inference time.

\subsection{Question Design Guidelines}

We design questions that test genuine 3D reasoning rather than linguistic priors. We follow three guiding principles:

\begin{itemize}
    \item \textbf{Rooted in entities.} The answer should be unambiguous and depend on a singular entity or a specific combination of entities, all of which are critical to the answer.
    \item \textbf{Limited ambiguity.} Questions should admit one correct answer, or a small set that always references the same critical entities. We adopt complementary metrics for cases where this cannot be enforced.
    \item \textbf{Tool use.} Questions should require one or more spatial or visual abstractions to answer accurately, ideally in combination. For example, asking about the color of a uniformly colored door is a poor question because most doors fall within a small distribution that an LLM can guess; asking about an unusually colored or transparent door requires a visual lookup.
\end{itemize}

\subsection{Question Categories}
We follow the taxonomy introduced in~\cite{sahoo2026conversationalimagesegmentationgrounding}.

\begin{enumerate}
    \item \textbf{Entities and Relations.} Identifying or referring to an entity through an open-ended description, possibly relative to other entities. \emph{Example:} which object is closest to the red plastic chair?
    \item \textbf{Affordance.} Reasoning about potential actions an agent can perform on an entity. \emph{Example:} where can I sit in this room?
    \item \textbf{Functionality.} Reasoning about the function of an entity. \emph{Example:} find me something that can help with cleaning the room.
    \item \textbf{Physics and Safety.} Reasoning about physical or safety constraints arising from entities and their relationships.
\end{enumerate} 

\subsection{Statistics}

The dataset contains 105 questions with 268 entities referenced in ScanNet++ scans. The average answer length is 282 characters and the average question length is 83 characters. 



\subsection{Example QA pairs}

\paragraph{Affordance with visuo-spatial reasoning.}
\textbf{Q.} You are a robot tasked with placing a large object somewhere stable and accessible for a human. Which surface in the room should I place it on? Eliminate all other surfaces with reasons.

\textbf{A.} You can use the \texttt{<component\_100>}table\texttt{</component\_100>} as it does not have any objects placed on top, unlike \texttt{<component\_200>}workbench\texttt{</component\_200>}, which has many objects placed on top, such as \texttt{<component\_300>}wrench\texttt{</component\_300>} and \texttt{<component\_300>}hammer\texttt{</component\_300>}.

\paragraph{Causal inference with implied entities and spatial reasoning.}
\textbf{Q.} I'm looking to clean my \texttt{<component\_18>}desk\texttt{</component\_18>}. Is there anything available in the room to do so, and if not, where should I look for more cleaning supplies in this room to supplement what is available?

\textbf{A.} You can use the \texttt{<component\_100>}tissues\texttt{</component\_100>} on top of the \texttt{<component\_200>}shelf\texttt{</component\_200>}. If you need more, you should look inside the \texttt{<component\_200>}shelf\texttt{</component\_200>} and \texttt{<component\_300>}cabinet\texttt{</component\_300>} under the \texttt{<component\_400>}desk\texttt{</component\_400>}.

\section{\benchmarkname{} Metrics}
\label{sec:benchmark-metrics}

We evaluate predictions on \benchmarkname{} along two complementary axes: \emph{response text evaluation}, which judges the textual rationale, and \emph{object grounding evaluation}, which judges whether the answer references the correct entities in the scene. Both are needed because \benchmarkname{} queries require a coherent multi-hop explanation and grounded references to specific components, and a method can succeed at one while failing at the other.

\paragraph{Why ScanQA-style metrics do not transfer.}
ScanQA reports n-gram overlap metrics (BLEU-4, ROUGE-L, CIDEr) and exactly match against short reference answers. 
Two properties of \benchmarkname{} make these metrics unsuitable. First, expected answers span multiple sentences, not single-line responses. 
Surface-level metrics penalize alternative correct phrasings: an answer that grounds the correct components in different textual prose receives a low score 
even when it is operationally identical to the reference. Second, ScanQA does not require explicit entity grounding, while \benchmarkname{} treats the 
predicted set of referenced components for evaluation. A correct answer that points at the wrong components is a different kind of 
failure from a less coherent answer that grounds on the right components, and a single score cannot tell these two cases apart.

\paragraph{Response text evaluation.}
We report two metrics on the textual rationale.

\begin{itemize}
    \item \textbf{AI-Judge.} An LLM is given the question, the expected answer, and the predicted answer, and produces an absolute score based on the semantic match between the expected and predicted answer. We use this as the primary text-quality metric because it tolerates paraphrase and accommodates the multi-step structure of expected answers.
    \item \textbf{METEOR.} We retain METEOR as a reproducible reference-based check that does not depend on a judge model. METEOR tolerates synonymy and reordering better than BLEU-4, which makes it a more useful reference signal for free-form answers.
\end{itemize}

\paragraph{Object grounding evaluation.}
The second axis treats the predicted answer as a set of grounded component identifiers. We extract the components wrapped in \texttt{<component\_ID>} tags from both the prediction and the expected answer, and compute precision, recall, and F1 over the two sets:

\begin{itemize}
    \item \textbf{Precision.} Fraction of components referenced in the prediction that also appear in the expected answer.
    \item \textbf{Recall.} Fraction of components in the expected answer that also appear in the prediction.
    \item \textbf{F1 Score.} Harmonic mean of precision and recall.
\end{itemize}

This evaluation captures whether the system identified the right entities, independent of the text. Combined with response text evaluation, it lets us separate grounding failures from reasoning failures.

\section{Additional Experimental Details and Results}
\label{sec:appendix_scanqa_comparison}

\begin{wrapfigure}{r}{0.5\textwidth}
    \centering
    \vspace{-10pt}
    \includegraphics[width=1\linewidth]{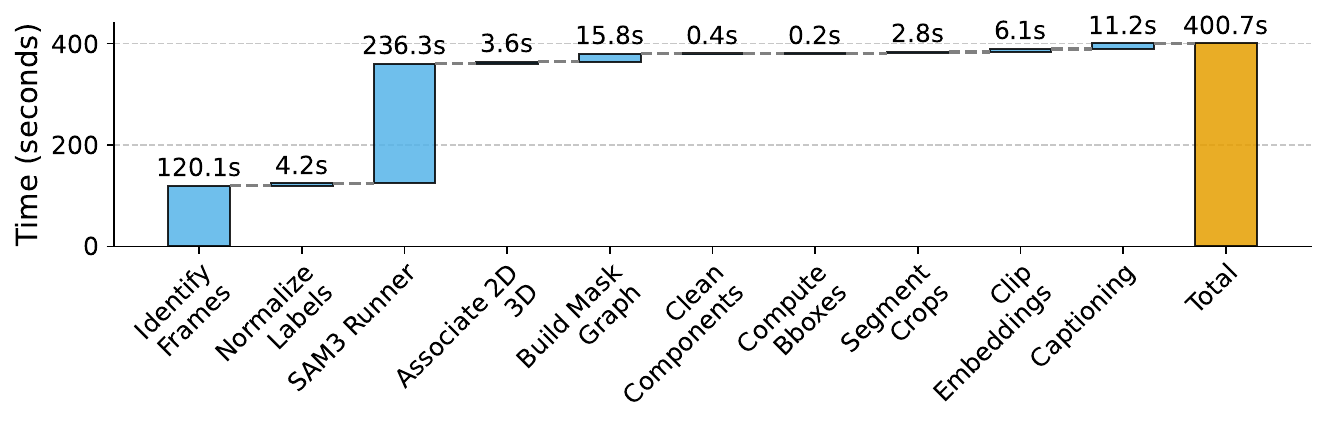}
    \caption{A waterfall chart detailing the runtime breakdown for creating the scene memory.}
    \label{fig:runtime_metrics}
    \vspace{-10pt}
\end{wrapfigure}

\subsection{Baselines}
To comprehensively evaluate \projectname{} on ScanQA, we compare against three distinct classes of baselines: \textbf{Expert models}~\citep{chen2020scanrefer,yu2019mcan,azuma2022scanqa,zhu20233dvista} that utilize dedicated 3D encoders and are trained with task-specific supervision for 3D question answering. \textbf{3D LMMs}~\citep{hong20233d_3dllm,chen2024ll3da,fu2024scenellm,huang2023chat3dv2,huang2024chatscene,zhu2024llava3d,huang2024leo,zheng2025video3dllm} that have undergone extensive pretraining or finetuning on large-scale 3D data. \textbf{Zero-shot methods}~\citep{li2024mvbench,bai2025qwen25vl,taguchi2025spatialprompting,zhang2024llavavideo} that operate without any task-specific 3D training. Finally, for our ablation studies designed to evaluate the impact of our spatial abstractions, we consider a ``No spatial tools'' configuration, where we only provide the agent access to external tools and not spatial tools.

\subsection{Extended ScanQA Comparison}

Table~\ref{tab:scanQAVal_full} presents the comprehensive evaluation results of \projectname{} on the ScanQA validation set, comparing against a wider range of expert models, 3D LMMs, and zero-shot methods.

\providecommand{\posdiff}[1]{\textcolor{green!45!black}{\tiny $\Delta$+#1}}
\providecommand{\negdiff}[1]{\textcolor{red!70!black}{\tiny $\Delta$#1}}

\begin{table}[h]
\centering
\renewcommand{\arraystretch}{1.12}
\resizebox{0.7\columnwidth}{!}{%
\begin{tabular}{lccccc}
\toprule
\textbf{Method}
& \textbf{CIDEr}$\uparrow$
& \textbf{BLEU-4}$\uparrow$
& \textbf{METEOR}$\uparrow$
& \textbf{ROUGE}$\uparrow$
& \textbf{EM}$\uparrow$ \\
\midrule

\multicolumn{6}{l}{\textit{\textbf{Expert models}}} \\
\midrule
ScanRefer+MCAN~\citep{chen2020scanrefer,yu2019mcan}
& 55.4 & 7.9 & 11.5 & 30.0 & 18.6 \\
ScanQA~\citep{azuma2022scanqa}
& 64.9 & 10.1 & 13.1 & 33.3 & 21.1 \\
3D-VisTA~\citep{zhu20233dvista}
& 69.6 & 10.4 & 13.9 & 35.7 & 22.4 \\

\midrule
\multicolumn{6}{l}{\textit{\textbf{3D LMMs}}} \\
\midrule
3D-LLM~\citep{hong20233d_3dllm}
& 69.4 & 12.0 & 14.5 & 35.7 & 20.5 \\
LL3DA~\citep{chen2024ll3da}
& 76.8 & 13.5 & 15.9 & 37.3 & -- \\
Scene-LLM~\citep{fu2024scenellm}
& 80.0 & 12.0 & 16.6 & 40.0 & 27.2 \\
Chat-3D v2~\citep{huang2023chat3dv2}
& 87.6 & 14.0 & -- & -- & -- \\
Chat-Scene~\citep{huang2024chatscene}
& 87.7 & 14.3 & 18.0 & 41.6 & 21.6 \\
LLaVA-3D~\citep{zhu2024llava3d}
& 91.7 & 14.5 & 20.7 & 50.1 & 27.0 \\
LEO~\citep{huang2024leo}
& 101.4 & 13.2 & 20.0 & 49.2 & 24.5 \\
Video-3D-LLM~\citep{zheng2025video3dllm}
& 102.1 & 16.2 & 19.8 & 49.0 & 30.1 \\
CVP~\citep{Chen_2026_WACV}
& \textbf{107.1} & 17.8 & 20.8 & \textbf{50.9} & \textbf{31.2} \\

\midrule
\multicolumn{6}{l}{\textit{\textbf{Zero-shot methods}}} \\
\midrule
VideoChat2~\citep{li2024mvbench}
& 49.2 \posdiff{29.24}
& 9.6 \posdiff{24.45}
& 9.5 \posdiff{24.54}
& 28.2 \posdiff{12.27}
& 19.2 \posdiff{7.24} \\

Qwen2.5-VL-7B~\citep{bai2025qwen25vl}
& 53.9 \posdiff{24.54}
& 3.0 \posdiff{31.05}
& 11.4 \posdiff{22.64}
& 29.3 \posdiff{11.17}
& -- \\

SpatialPrompting~\citep{taguchi2025spatialprompting}
& 87.69 \negdiff{-9.25}
& --
& 16.85 \posdiff{17.19}
& 43.39 \negdiff{-2.92}
& 27.34 \negdiff{-0.90} \\

LLaVA-Video~\citep{zhang2024llavavideo}
& 88.7 \negdiff{-10.26}
& 3.1 \posdiff{30.95}
& 17.7 \posdiff{16.34}
& 44.6 \negdiff{-4.13}
& -- \\

\midrule
\textbf{Ours}
& 78.44
& \textbf{34.05}
& \textbf{34.04}
& 40.47
& 26.44 \\

\bottomrule
\end{tabular}}
\vspace{4pt}
\caption{Comprehensive performance on the ScanQA validation set. All metrics are higher-is-better. In the zero-shot section, the smaller text next to the numbers reports the absolute difference between \projectname{} and the corresponding baseline, computed as \projectname{} minus the baseline.}
\label{tab:scanQAVal_full}
\vspace{-10pt}
\end{table}

\subsection{Microbenchmarks}
\label{subsec:microbenchmarks}
Figure~\ref{fig:runtime_metrics} illustrates the runtime metrics associated with creating the scene memory from raw posed RGB-D data. The waterfall chart breaks down the execution time across various processing stages each of which are described in Appendix~\ref{sec:appendix-instance-seg}.


\newpage
\clearpage

\end{document}